\documentclass[10pt,twocolumn,letterpaper]{article}
\usepackage[pagenumbers]{cvpr}

\usepackage{cuted} 

\usepackage{currfile} 

\usepackage{caption} 

\usepackage{multirow}
\usepackage{enumitem}
\usepackage{tikz}
\usetikzlibrary{arrows.meta,positioning}
\definecolor{cvprblue}{rgb}{0.21,0.49,0.74}
\usepackage[pagebackref,breaklinks,colorlinks,allcolors=cvprblue]{hyperref}

\def\paperID{}
\def\confName{arXiv}
\def\confYear{2026}

\newcommand{\method}{HandsOnWorld\xspace}

\newcommand{\vd}{\mathbf{d}}
\newcommand{\vo}{\mathbf{o}}
\newcommand{\vp}{\mathbf{p}}
\newcommand{\vn}{\mathbf{n}}
\newcommand{\vell}{\boldsymbol{\ell}}
\newcommand{\RR}{\mathbb{R}}
\newcommand{\lhand}{\vell_{\mathrm{hand}}}
\newcommand{\lcam}{\vell_{\mathrm{cam}}}

\title{HandsOnWorld: Unconstrained Egocentric Video Generation with
Camera-Disentangled Hand Control}

\author{%
Yushuo Chen$^{1,2}$ \quad
Xiaoyu Shi$^{2}$ \quad
Xiaoshi Wu$^{3}$ \quad
Xintao Wang$^{2}$ \quad
Pengfei Wan$^{2}$ \quad
Yebin Liu$^{1}$\\[0.3em]
$^{1}$Tsinghua University \quad
$^{2}$Kling Team, Kuaishou Technology \quad
$^{3}$Chinese University of Hong Kong
}

\begin{document}
\maketitle

\begin{strip}
  \centering
  \includegraphics[width=\textwidth]{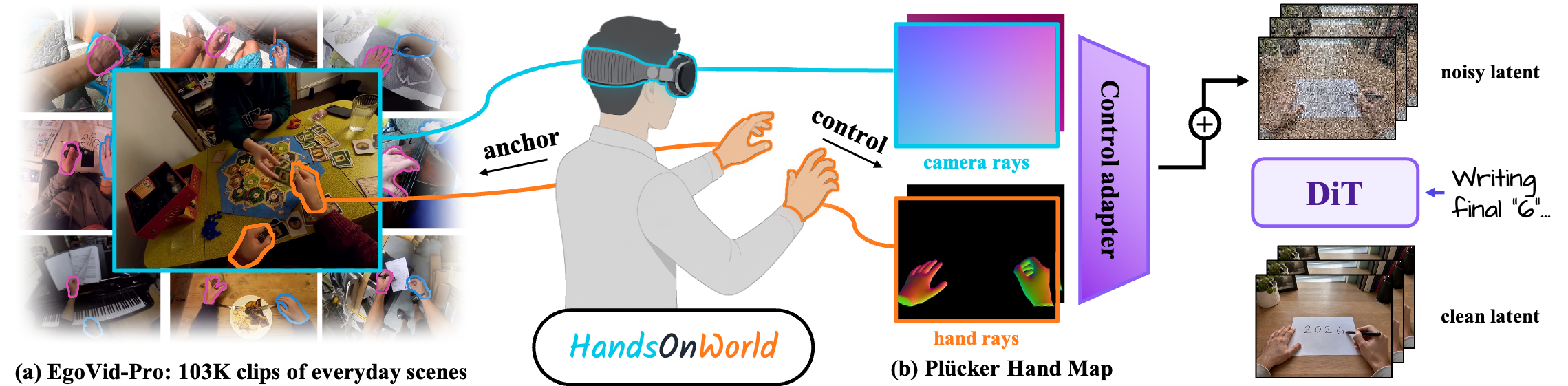}
  \captionof{figure}{\textbf{Overview of HandsOnWorld.}
    We propose (a) the protagonist-centric dataset \textbf{EgoVid-Pro} and
    (b) the \textbf{Pl\"{u}cker Hand Map}, a camera-disentangled hand control
    representation, enabling accurate 3D hand control and plausible
    interactions in everyday scenes.}
  \label{fig:teaser}
\end{strip}

\begin{abstract}
We present \method, a framework for hand-controlled egocentric video
generation that learns directly from unconstrained monocular video.
Prior generators depend on 3D hand annotations from
multi-view or marker-based motion capture, confining them to narrow,
instrumented scene distributions.
To bridge this gap, we introduce a \emph{protagonist-centered annotation
pipeline} that filters monocular 3D reconstructions at the
action-semantic, image-quality, and 3D-geometric levels, yielding
\textbf{EgoVid-Pro}, a dataset of clean, protagonist-only hand
trajectories spanning 103K clips and roughly 12M frames across diverse
everyday scenes.
These unconstrained scenes exhibit substantial camera ego-motion that
is largely absent from tabletop captures, exposing the entanglement of
camera and hand motion in existing camera-space control signals.
We therefore propose the \textbf{Pl\"{u}cker Hand Map}, which extends
Pl\"{u}cker rays from camera geometry to the hand surface, representing
hand motion in the same world frame as the camera and disentangling
the two motion sources at the representation level.
Experiments show that \method outperforms prior methods in visual fidelity and control
accuracy and generalizes beyond lab settings.
\end{abstract}


\section{Introduction}

Imagine reaching out to pick up a coffee mug in a mountain cabin, flipping
cards on a cluttered kitchen table, or sketching on paper in a sunlit studio.
Egocentric video generation is approaching this kind of immersive realism, but
only if we can faithfully control what our hands do.
Recent generative video
models~\citep{wang2025wan,openai2024sora,yang2024cogvideox} produce
photorealistic footage from text or image prompts, and increasingly serve as
interactive world simulators that predict how a scene unfolds in
response to an agent's actions.
One line of work explores navigation in game and driving
environments~\citep{bruce2024genie,valevski2024gamengen,hu2023gaia1,gao2024vista};
a parallel line brings world simulators to human embodiment, conditioning
generation on body or hand pose to simulate egocentric
interaction~\citep{tu2025playerone,bai2025peva}.
Hands are our primary interface with the physical world, and fine-grained
hand control unlocks a manipulable form of egocentric generation:
experiencing the generated world through one's own hands, free from
the constraints of any particular environment.

\begin{figure*}
  \centering
  \includegraphics[width=\textwidth]{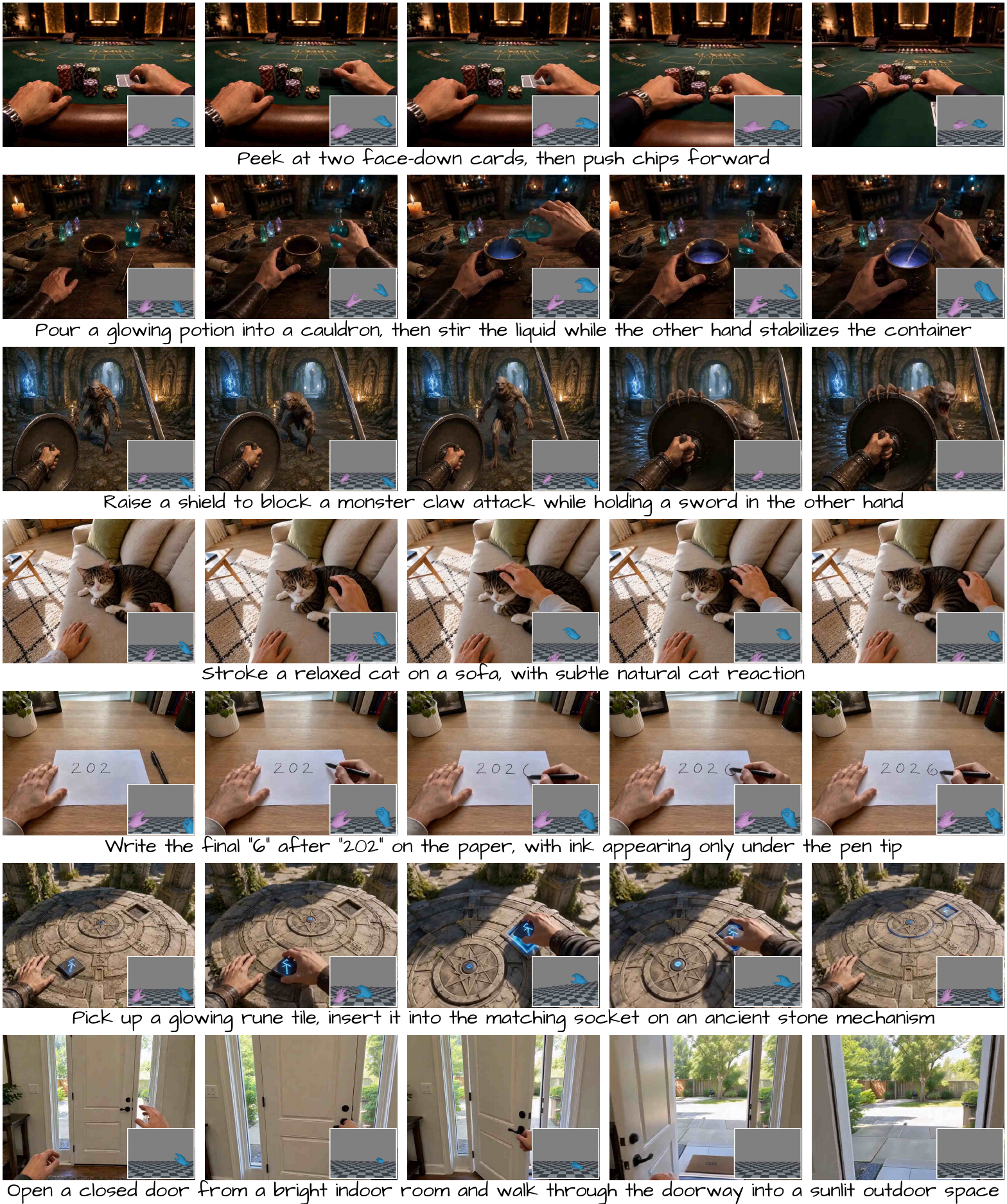}
  \caption{\textbf{HandsOnWorld: Unconstrained 3D hand-controlled egocentric video generation.}
    Given the first frame and a target 3D camera and hand trajectory, our
    method synthesizes temporally coherent egocentric interactions across
    diverse everyday scenes, objects, and actions, generalizing far beyond the
    controlled tabletop settings of prior work.
    The first frames are generated with GPT-Image-2, and input text prompts are
    augmented before being passed to the video model.}
  \label{fig:showcases}
\end{figure*}

Achieving this generality is fundamentally constrained by training data.
Acquiring accurate 3D hand pose has required calibrated multi-camera rigs,
confining hand supervision to controlled, instrumented capture.
The resulting datasets form a data annotation pyramid where the
fidelity of hand supervision is inversely coupled to scene diversity.
At the base, large-scale in-the-wild egocentric
corpora~\citep{grauman2022ego4d,ma2024nymeria,lv2024aea} capture unconstrained
everyday scenes, but provide only coarse body pose without finger-level
articulation.
At the intermediate level, recordings collected at fixed multi-camera sites
recover 3D hand pose through labor-intensive manual annotation, yet cover only
a small set of staged activities~\citep{grauman2024egoexo4d,ohkawa2023assemblyhands}.
At the apex, dense multi-camera capture and instrumented headsets yield highly
precise hand-object
annotations~\citep{taheri2020grab,chao2021dexycb,kwon2021h2o,liu2022hoi4d,fan2023arctic,liu2024taco,zhan2024oakink2,banerjee2025hot3d,hoque2025egodex},
but confine capture to a fixed tabletop.
Recent related
methods~\citep{wang2026hand2world,li2026egohoi,xie2026genreality,zhang2026occlusionhand}
are trained on these annotated tiers and inherit their restricted scene
diversity.


Recent advances in monocular 3D understanding offer a path around this
bottleneck.
Feedforward models now directly predict camera parameters, dense scene
geometry, or 3D object structure from a handful of RGB
inputs~\citep{wang2024dust3r,leroy2024mast3r,wang2025moge,wang2025vggt,sam3d2025};
an analogous line of work has matured for hands, recovering world-space 3D
hand trajectories from ordinary monocular
video~\citep{pavlakos2024hamer,potamias2024wilor,yu2025dynhamr,zhang2025hawor}.
Building on these priors, we annotate
EgoVid-5M~\citep{wang2024egovid5m}, a curated subset of
Ego4D~\citep{grauman2022ego4d}, using only monocular reconstruction.
However, the unconstrained nature of everyday egocentric scenes surfaces two
challenges that controlled lab data does not face.
First, \textbf{protagonist hand identification}: in everyday scenes, off-the-shelf
hand reconstruction returns trajectories from any visible hand, including
bystanders, hand-like false positives, and unstable detections under motion
blur or occlusion. The protagonist's hands must be isolated from this noise
to form clean training data.
Beyond annotation, a second challenge arises in how the hand is represented
during conditioning. \textbf{Camera-hand motion entanglement}: unconstrained
egocentric scenes are dominated by substantial camera ego-motion that is
largely absent in tabletop captures. Existing camera-space control signals,
such as projected 2D joints or rendered mesh images, encode only the
camera-relative pose of the hand, so equivalent signals can correspond to very
different absolute 3D motions, making the hand's true 3D trajectory ambiguous.

To address these two challenges, we propose two complementary solutions.
First, we construct \textbf{EgoVid-Pro}, a large-scale egocentric dataset
of clean, protagonist-only 3D hand annotations, built by a
\textbf{protagonist-centered annotation pipeline} that filters in-the-wild
detections at the semantic, image, and 3D-geometry levels.
Second, we propose the \textbf{Pl\"{u}cker Hand Map}, a 3D-aware control
signal that extends the Pl\"{u}cker-ray parameterization of camera
pose~\citep{he2024cameractrl} from camera rays to the hand surface.
Representing the hand in the same world frame as the camera disentangles its absolute 3D motion from camera ego-motion. This absolute placement determines whether the generated hand actually reaches the object in the world, enabling more accurate control and generating more physically plausible interactions.

Overall, our contributions are:

\begin{itemize}

  \item \textbf{Unconstrained egocentric hand-controlled video generation.}
  We propose a 3D hand-controlled egocentric video generation framework that
  does not rely on multi-view or marker-based motion capture, enabling
  training on unconstrained monocular video and generalization to diverse
  everyday scenes.

  \item \textbf{EgoVid-Pro dataset.}
  By applying a protagonist-centered annotation pipeline to large-scale
  in-the-wild egocentric video, we curate EgoVid-Pro, a dataset of clean,
  protagonist-only 3D hand trajectories that matches the largest existing
  3D-hand-annotated egocentric dataset in scale while spanning far more
  diverse, everyday scenes.

  \item \textbf{Pl\"{u}cker Hand Map.}
  A unified world-space control signal pairing camera Pl\"{u}cker rays with
  surface-normal rays, disentangling the absolute 3D hand motion from camera ego-motion at the representation level.

\end{itemize}

\section{Related Work}

\paragraph{Controllable video generation.}
Controllable video generation extends diffusion video
models~\citep{blattmann2023videoldm,yang2024cogvideox,openai2024sora,kong2024hunyuanvideo,wang2025wan,nvidia2025cosmos}
with conditioning beyond text or image prompts.
One family conditions on 2D image signals (Canny edges, normal maps,
drag points, human pose), supporting tasks from character animation to
controllable video
editing~\citep{yin2023dragnuwa,wu2024draganything,zhang2025tora,hu2024animateanyone,zhu2024champ,shao2024human4dit,jiang2025vace}.
Others pursue 3D-aware control through camera pose, ranging from
learned motion modules~\citep{guo2024animatediff} and independent
camera/object trajectories~\citep{wang2024motionctrl} to per-pixel
Pl\"{u}cker-ray representations~\citep{he2024cameractrl} and camera-trajectory
replay~\citep{bai2025recammaster}; a few push further to full 3D
object pose, specifying 6D camera-and-object
trajectories~\citep{shuai2025fmc,fu2025trajmaster}.

\paragraph{Egocentric world simulator.}
Egocentric world simulators model how a scene unfolds from the agent's own
viewpoint in response to its actions, along an axis of increasingly
fine-grained control.
At the coarse end, early systems drive generation with abstract inputs such
as button presses, targeting synthetic
game~\citep{bruce2024genie,valevski2024gamengen} and
driving~\citep{hu2023gaia1,gao2024vista} environments.
A complementary thread grounds control in human embodiment instead:
PlayerOne~\citep{tu2025playerone} and PEVA~\citep{bai2025peva} drive
generation from SMPL or whole-body kinematics, but capture only coarse,
large-scale motion without finger articulation.
The finest-grained thread targets hand pose directly, injecting 3D hand pose
into the diffusion backbone: Hand2World~\citep{wang2026hand2world} projects
an occlusion-invariant MANO mesh, GenReality~\citep{xie2026genreality} and
EgoHOI~\citep{li2026egohoi} tokenize 3D skeletons or hand-mesh embeddings.
SpriteHand~\citep{li2025spritehand} instead edits existing footage in a
video-to-video HOI setting rather than generating from scratch.
These generation-based methods mostly train on controlled lab
captures~\citep{banerjee2025hot3d,fan2023arctic,liu2022hoi4d} of limited
scale; \citet{zhang2026occlusionhand}, denoted JointControl for brevity, and
EgoSim~\citep{hao2026egosim} subsequently scale up hand annotation for
robotic-manipulation tasks, yet their data remains confined to tabletop,
largely static, single-operator scenes that fail to generalize to
unconstrained everyday activity.
Moreover, all existing works represent the hand control signal in camera space,
entangling camera ego-motion with hand motion and preventing accurate
generation of the hand's interaction with the environment.

\paragraph{3D hand reconstruction.}
Our annotation pipeline depends on accurate 3D hand reconstruction, built on
parametric models such as MANO~\citep{romero2017mano},
SMPL~\citep{loper2015smpl}, and SMPL-X~\citep{pavlakos2019expressive}, whose
low-dimensional pose spaces underlie both our pipeline and our control
signal.
The most precise annotations come from multi-view capture: marker-based
optical mocap triangulates and fits a parametric
model~\citep{taheri2020grab,fan2023arctic,zhan2024oakink2,banerjee2025hot3d},
markerless RGB-D relies on offline keypoint
optimization~\citep{chao2021dexycb,liu2024taco}, and the same multi-camera
principle drives real-time VR hand
tracking~\citep{han2020megatrack,han2022umetrack}. These methods require
instrumentation that confines capture to controlled indoor settings.
Monocular reconstruction lifts this constraint: HaMeR~\citep{pavlakos2024hamer}
scales transformer-based recovery for strong in-the-wild generalization, and
WiLoR~\citep{potamias2024wilor} adds a real-time detector for robust
multi-hand recovery.
Recent work extends this to 4D recovery under a moving camera:
Dyn-HaMR~\citep{yu2025dynhamr} jointly optimizes per-frame MANO with the
camera trajectory, and HaWoR~\citep{zhang2025hawor} couples MANO with
egocentric SLAM for world-space trajectories.
Our pipeline builds on HaWoR with WiLoR detection as the front-end,
addressing the additional challenges of in-the-wild egocentric scenes.

\section{Overview and Preliminaries}
\label{sec:overview}

\method enables unconstrained egocentric video generation with fine-grained
3D hand control by leveraging large-scale in-the-wild video (Figure~\ref{fig:teaser}).
Specifically,we introduce a \textbf{protagonist-centered annotation pipeline}
(Section~\ref{sec:data}) that filters everyday footage and recovers
high-fidelity annotations of the protagonist's actions, yielding
\textbf{EgoVid-Pro} corpus.
As such data exhibits considerably more complex head-hand motion than
controlled captures, we further introduce \textbf{Pl\"{u}cker Hand Map}
(Section~\ref{sec:method_signal}), a novel control representation that disentangles hand
motion from camera ego-motion for more accurate controlled generation.

\paragraph{Wan video diffusion model.}
\method builds on the 5B and 14B variants of Wan2.2-I2V~\citep{wang2025wan},
which adopt a Diffusion Transformer (DiT) operating in a compressed video
latent space. The 5B model conditions on the first frame by concatenating its
latent with the noisy video latent, whereas the 14B model uses dedicated
cross-attention for first-frame conditioning. The 14B model further adopts a
two-stage denoising schedule that separates a high-noise stage (coarse
structure) from a low-noise stage (fine details), improving temporal
coherence on long sequences.
The model is trained with the standard diffusion denoising
objective~\citep{ho2020ddpm}.
Given a clean video latent $\mathbf{x}_0$, the forward process gradually adds
Gaussian noise,
\begin{equation}
\mathbf{x}_t = \sqrt{\bar{\alpha}_t}\,\mathbf{x}_0
   + \sqrt{1-\bar{\alpha}_t}\,\boldsymbol{\epsilon},
\qquad \boldsymbol{\epsilon} \sim \mathcal{N}(\mathbf{0}, \mathbf{I}),
\end{equation}
where $t \in \{1,\dots,T\}$ indexes diffusion timesteps and
$\bar{\alpha}_t = \prod_{s\le t}(1-\beta_s)$ is a monotonically decreasing
noise schedule.
A denoising network $\boldsymbol{\epsilon}_\phi$, conditioned on inputs
$\mathbf{c}$ such as text or the first frame, is trained to predict the added
noise via
\begin{equation}
\mathcal{L}_{\text{diff}} = \mathbb{E}_{t,\mathbf{x}_0,\boldsymbol{\epsilon}}
\bigl\|\boldsymbol{\epsilon} - \boldsymbol{\epsilon}_\phi(\mathbf{x}_t, t, \mathbf{c})\bigr\|^2.
\end{equation}
At inference, samples are drawn by iteratively denoising
$\mathbf{x}_T \sim \mathcal{N}(\mathbf{0}, \mathbf{I})$.

\paragraph{MANO.}
MANO~\citep{romero2017mano} is a parametric hand model mapping pose
$\boldsymbol{\theta} \in \RR^{48}$ and shape $\boldsymbol{\beta} \in \RR^{10}$
to a mesh of $|\mathcal{V}|=778$ vertices articulated by $K=16$ joints.
MANO uses pose- and shape-dependent corrective blend shapes followed by
linear blend skinning.
The related SMPL body model~\citep{loper2015smpl} provides a full-body
kinematic chain spanning head, torso, arms, and wrists, which we leverage
to identify the protagonist's hands in unconstrained footage by enforcing
kinematic consistency between detected wrists and the egocentric viewpoint
(Section~\ref{sec:data_3d}).

\paragraph{HaWoR.}
HaWoR~\citep{zhang2025hawor} reconstructs world-space hand motion from
monocular egocentric video.
It first detects and tracks hands across frames, yielding tracklets:
temporally linked detections of a single hand, each labeled by handedness
and scored with a per-frame confidence.
A transformer then regresses per-frame camera-space MANO parameters for
each tracklet, while an adaptive SLAM module estimates the world-space
camera trajectory; combining the two lifts each tracklet to world-space
motion $\{\mathcal{M}_t\}_{t=1}^{T}$.

\begin{figure*}[t]
  \centering
  \includegraphics[width=\linewidth]{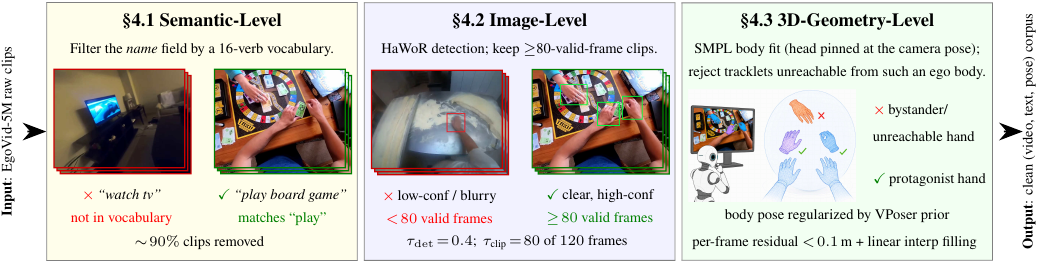}
  \caption{Overview of the protagonist-centered annotation pipeline.
    Starting from EgoVid-5M, we progressively discard clips that fail
    semantic, image-quality, or 3D-geometry criteria,
    yielding clean (video, text, human pose) tuples for training.}
  \label{fig:pipeline}
\end{figure*}

\section{Data Annotation Pipeline}
\label{sec:data}

Scaling egocentric hand-controlled generation requires clean,
protagonist-centered hand annotations recovered from unconstrained footage.
Starting from EgoVid-5M~\citep{wang2024egovid5m}, a curated, pre-segmented
subset of Ego4D with textual action labels~\citep{grauman2022ego4d}, our
\textbf{protagonist-centered annotation pipeline} applies three progressive
filtering stages (semantic, image, and 3D geometric) to produce
\textbf{EgoVid-Pro}, a training corpus of (video, text, hand-trajectory)
triples in which the protagonist is consistently the agent of the depicted
hand action (Figure~\ref{fig:pipeline}).

\subsection{Semantic-Level Filtering}

EgoVid-5M provides two text annotations per clip: a short \emph{name}
summarizing the protagonist's activity (\eg, \textit{``engage with
phone''}, \textit{``paint wall''}), and a longer \emph{LLaVA caption}
describing the full visual content.
We use the name for filtering and retain the LLaVA caption as the
text-conditioning signal during generator training.
A large fraction of names refer to passive or socially interactive
activities (\eg, \textit{engage}, \textit{watch}) that
carry little hand-manipulation signal.
We curate a vocabulary of \textbf{16} \emph{action verbs} corresponding to
concrete hand-driven manipulations (\eg, \textit{paint}, \textit{move},
\textit{grab}), and keep only clips whose
name contains at least one such verb.
This stage removes roughly $90\%$ of the corpus while retaining the most
manipulation-rich footage.

\subsection{Image-Level Filtering}

We apply HaWoR~\citep{zhang2025hawor} to the surviving clips.
HaWoR's detect-and-track stage emits a confidence score for each detected
bounding box; under rapid camera motion or severe occlusion, low-confidence
detections produce spurious tracks that degrade annotation quality.
We tighten the detection threshold from HaWoR's default $0.2$ to
$\tau_{\det} = 0.4$, treating any lower-confidence frame as a missed
detection. Then we discard any clip with fewer
than $\tau_{\text{clip}} = 80$ retained detections from $120$ frames, removing sequences
that no longer carry enough usable hand observations.


\subsection{3D-Geometry-Level Filtering: Identifying the Protagonist's Hands}
\label{sec:data_3d}

\begin{figure}[t]
  \centering
  \newcommand{\failimgheight}{0.30\linewidth}
  \tabcolsep=0pt
  \begin{tabular}{@{}c@{\hspace{3pt}}c@{}}
    \multicolumn{2}{c}{\scriptsize\textbf{Image-plane orientation heuristic}}\\[-1pt]
    \includegraphics[height=\failimgheight]{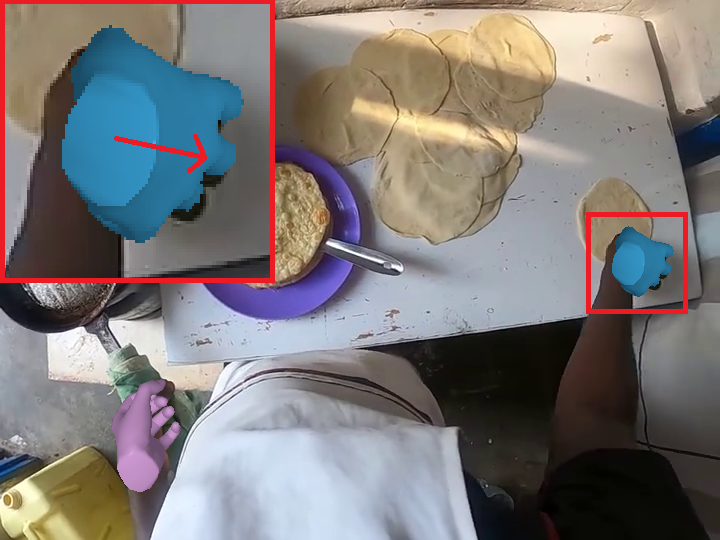}
    &
    \includegraphics[height=\failimgheight]{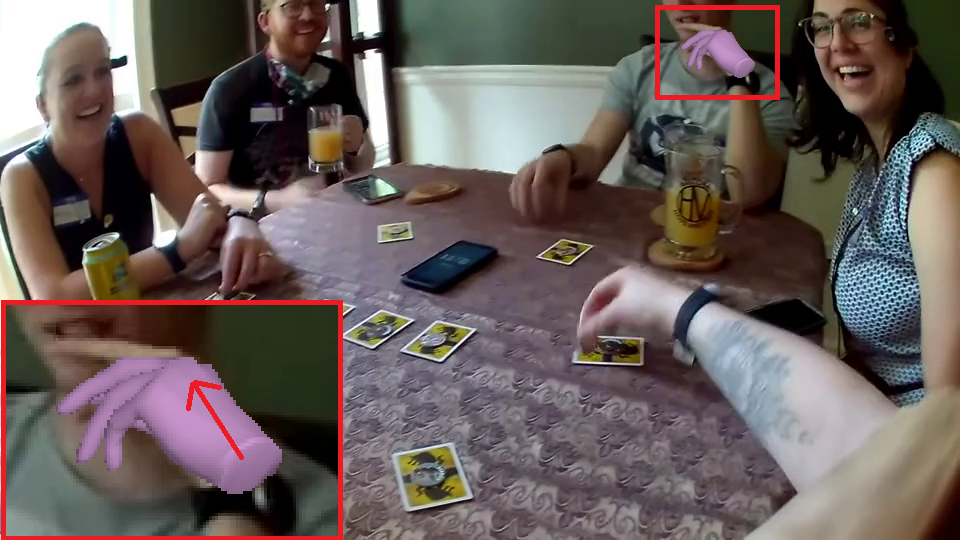}
    \\[-1pt]
    \parbox{0.38\linewidth}{\centering\scriptsize \textbf{False reject:} protagonist hand points downward}
    &
    \parbox{0.52\linewidth}{\centering\scriptsize \textbf{False retain:} bystander hand points upward}
    \\[5pt]
    \multicolumn{2}{c}{\scriptsize\textbf{Camera-facing direction heuristic}}\\[-1pt]
    \includegraphics[height=\failimgheight]{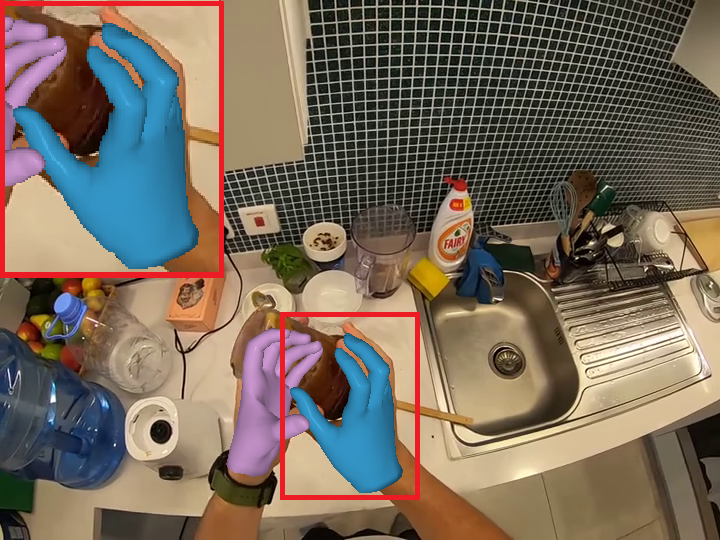}
    &
    \includegraphics[height=\failimgheight]{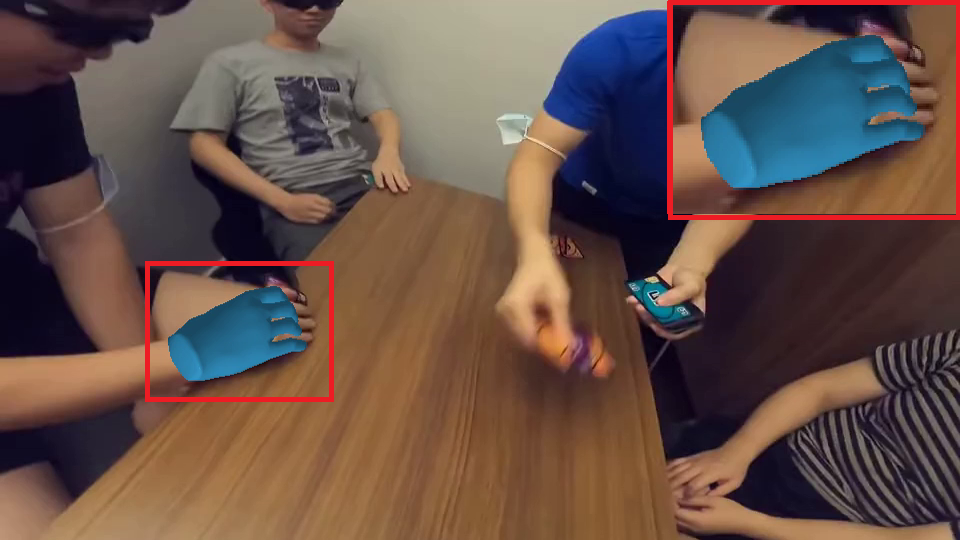}
    \\[-1pt]
    \parbox{0.38\linewidth}{\centering\scriptsize \textbf{False reject:} valid hands point towards the camera}
    &
    \parbox{0.52\linewidth}{\centering\scriptsize \textbf{False retain:} bystander hand points outward the camera}
  \end{tabular}
  \caption{\textbf{Failure cases of naive camera-space heuristics.} We illustrate two
    vanilla rules based on the hand orientation. \textbf{(top)} Image-plane
    orientation, expecting the protagonist's hand to point upward.
    \textbf{(bottom)} Camera-facing direction, expecting alignment with
    the optical axis. Two columns show false
    rejections of protagonist hands and false retention of
    bystander hands, respectively.}
  \label{fig:naive_failure}
\end{figure}

After semantic and image filtering, a non-trivial fraction of the
surviving detections still correspond to bystander hands or hand-like
objects.
Simple camera-space heuristics, such as thresholding the hand orientation (the vector from the hand root to the
middle-finger MCP joint) in
the image plane or against the optical axis, cannot reliably separate
them: valid protagonist hands appear in non-canonical poses, while
bystander hands and false detections occupy the same orientation modes
(Figure~\ref{fig:naive_failure}).

We instead recast the test as a \textbf{world-space body-fit problem}.
For each clip, we jointly fit a single SMPL~\citep{loper2015smpl} body to
all estimated hand tracklets.
The body's head is anchored to the egocentric camera, and its wrist and
middle-finger joints are pulled toward the per-frame HaWoR anchors.
We share one body shape across all tracklets in a clip and regularize the
articulated pose with the VPoser~\citep{pavlakos2019expressive} prior.
A tracklet survives only if this single first-person body can reach its
hand anchors at every observed frame;
unreachable tracklets, including bystander hands and unstable detections, are
rejected.
The full objective and hyperparameters are provided in
\cref{sec:supp_smpl_fit}.
Undetected frames are labeled by linearly interpolating the MANO parameters between the bracketing detections, applied only when the poses of the two endpoints are sufficiently similar.

\begin{figure}[t]
  \centering
  \includegraphics[width=0.98\linewidth]{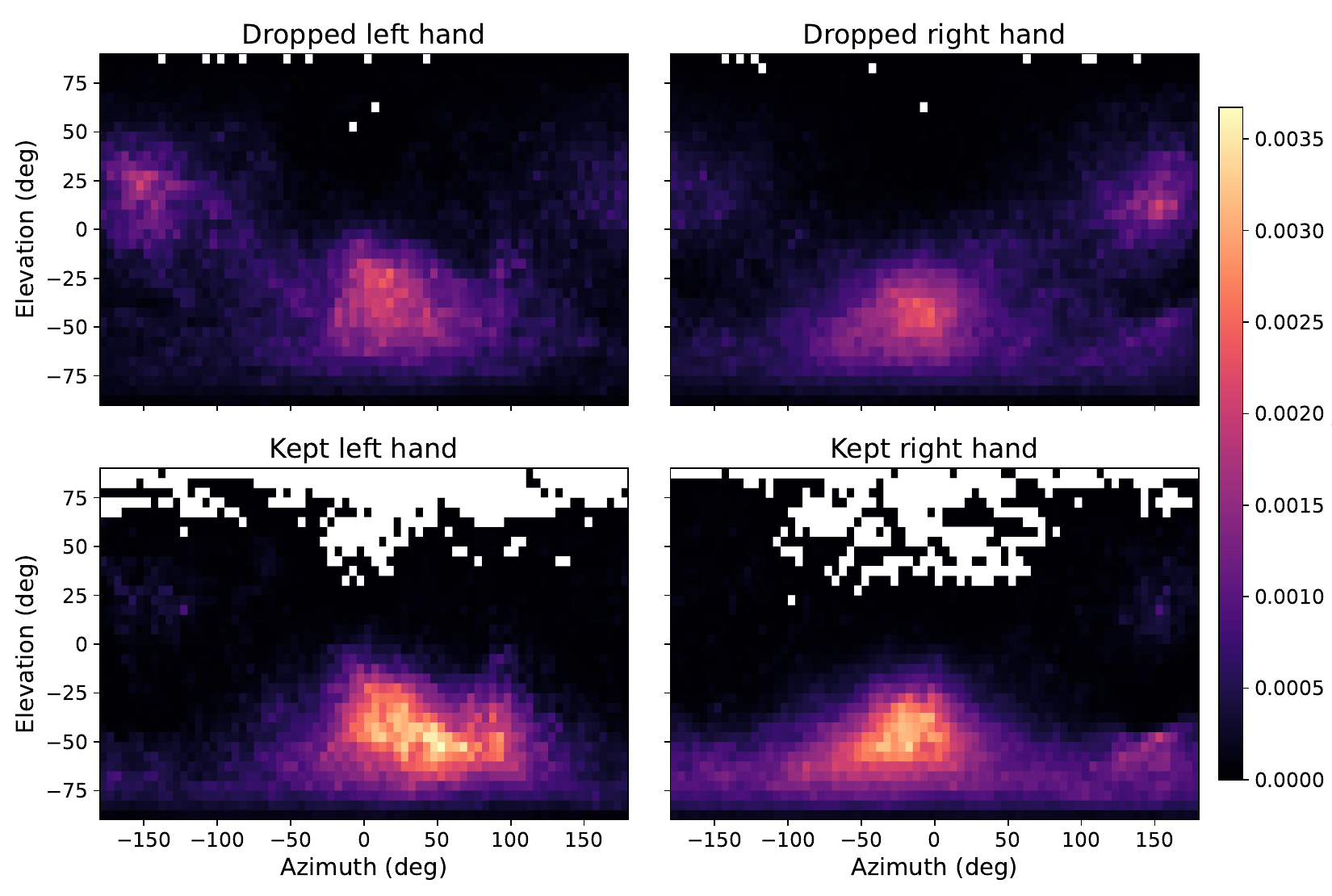}
  \caption{\textbf{Camera-space hand orientation statistics for tracklets retained
    and rejected by our 3D-geometry filter.} Azimuth and elevation are the horizontal and vertical angles of the hand orientation in the camera frame. 
    Although retained hands follow
    the expected egocentric bias, rejected tracklets overlap these regions,
    confirming that fixed camera-space thresholds are insufficient for
    robust identification.}
  \label{fig:smpl_stats}
\end{figure}

\Cref{fig:smpl_stats} compares the camera-space hand-orientation
distributions of tracklets retained and rejected by this filter across
the full corpus.
Retained hands follow the expected egocentric bias, but rejected
tracklets overlap the same orientation modes, confirming that no fixed
camera-space threshold separates the two and that a 3D-based
filter is necessary.

\section{3D-Aware Hand Control Signal}
\label{sec:method_signal}


In this section, we introduce the \textbf{Pl\"{u}cker Hand Map}, which extends
the world-space Pl\"{u}cker-ray
parameterization~\citep{he2024cameractrl} from camera rays to rays bound
to the hand surface, providing a representation that disentangles hand
motion from camera ego-motion at the input level.
The map is then passed through a lightweight convolutional encoder and added
(as a residual) to the noisy video latent at each denoising step, analogous
to ControlNet-style injection~\citep{zhang2023controlnet}.

\subsection{Pl\"{u}cker-Ray Representation}
\label{sec:plucker}

\paragraph{Camera ray.}
Following prior camera-control work~\citep{he2024cameractrl}, we
parameterize each pixel $(u,v)$ in frame $t$ by its world-space camera ray
$\lcam^{u,v,t} \in \RR^6$:
\begin{equation}
  \lcam^{u,v,t} = \bigl(\vd^{u,v,t},\; \vo^t \times \vd^{u,v,t}\bigr),
  \label{eq:cam_plucker}
\end{equation}
where $\vo^t$ is the camera center in world space at time $t$, and
$\vd^{u,v,t} = \mathbf{R}_t \mathbf{K}^{-1}[u,v,1]^\top /
\|\mathbf{R}_t \mathbf{K}^{-1}[u,v,1]^\top\|$
is the unit ray direction in world coordinates, computed from the camera
intrinsics $\mathbf{K}$ and extrinsic rotations $\mathbf{R}_t$.

\paragraph{Hand surface-normal ray.}
We rasterize the posed MANO mesh $\mathcal{M}_t$ into the camera frame
using nvdiffrast~\citep{laine2020nvdiffrast}; for each pixel $(u,v)$ whose
camera ray intersects the mesh, let $\vp^{u,v,t} \in \RR^3$ be the
world-space intersection point and $\vn^{u,v,t}$ the outward surface
normal at that point.
We define the hand surface-normal ray:
\begin{equation}
  \lhand^{u,v,t} = \bigl(\vn^{u,v,t},\; \vp^{u,v,t} \times \vn^{u,v,t}\bigr).
  \label{eq:hand_plucker}
\end{equation}
For pixels not covered by the hand, we set $\lhand^{u,v,t} = \mathbf{0}
\in \RR^6$.

\paragraph{Combined control map.}
The per-frame Pl\"{u}cker Hand Map concatenates the camera and hand rays into a
12-channel map $\mathbf{f}^t \in \RR^{H \times W \times 12}$:
Keeping the camera and hand rays in separate channels lets the user supply
any combination of camera and hand trajectories as input.
\begin{equation}
  \mathbf{f}^{u,v,t} = \bigl[\lcam^{u,v,t};\, \lhand^{u,v,t}\bigr].
  \label{eq:combined}
\end{equation}

\subsection{Disentanglement Property}
\label{sec:disentangle}


Writing $\lhand(P) = (\vn_P, \vp_P \times \vn_P)$ for the Pl\"{u}cker ray
attached to a mesh point $P$ with world-space position $\vp_P$ and
normal $\vn_P$, the value $\lhand^{u,v,t}$ at pixel $(u,v)$ is exactly
$\lhand(P_{u,v,t})$ where $P_{u,v,t}$ is the mesh point hit by that
pixel's camera ray.
Camera motion changes which mesh point a pixel hits, but the ray attached
to any fixed mesh point depends only on the hand pose.
Tracking a single mesh point $P$ under pure translation, we examine both the
pixel it projects to and the value $\lhand(P)$ stored there:
\begin{itemize}
  \item[(i)] \textbf{Camera-only translation}
    ($\vo^t \to \vo^t + \Delta$):
    $P$ projects to a different pixel, whose homogeneous image coordinate
    shifts by $-K \mathbf{R}_t^\top \Delta$; the hand, however, stays static in the world, hence $\lhand(P)$ remains unchanged.
  \item[(ii)] \textbf{Camera and hand translate together by $\Delta$}
    ($\vo^t \to \vo^t + \Delta$, $\vp_P \to \vp_P + \Delta$):
    $P$ projects to the same pixel, but $\vp_P$ shifts by $\Delta$, so
    the moment of $\lhand(P)$ shifts by $\Delta \times \vn_P$.
\end{itemize}

These two cases expose the fundamental advantage of world-space hand
representations over their camera-space counterparts (e.g., rendered meshes, or
depth maps).
A camera-space signal is a function of the camera-to-hand relative
pose alone; it therefore cannot separate the two motion sources: it varies in
case~(i), where the hand is stationary in the world, yet remains constant in
case~(ii), where the hand undergoes genuine world-space translation.
Our surface-normal ray exhibits the converse behavior, remaining invariant to
camera ego-motion while varying by $\Delta \times \vn_P$ under hand motion
alone.
The two motions are thereby disentangled at the level of the control signal.

\paragraph{Normal-degeneration corner case.}                                                                         
When the outward normal $\vn$ is nearly parallel to the ray direction $\vd$                                          
(\ie, the camera looks exactly along the surface normal), the moment                                                 
$\vp \times \vn$ becomes numerically sensitive.                                                                      
In practice, this grazing configuration is rare for typical hand-to-camera                                           
geometries and we have not observed instabilities in training.     




\begin{table*}[t]
  \centering
  \caption{\textbf{Quantitative comparison of different training-data conditions.}
    Training on our annotated EgoVid-Pro dataset avoids the data scale-label tradeoff and achieves the best performance on every metric. Best per column in \textbf{bold}.
    }
  \label{tab:data_effectiveness}
  \tabcolsep=3pt
  \small
  \begin{tabular}{lccccccccccc}
    \toprule
    & \multicolumn{6}{c}{Visual Quality} & \multicolumn{2}{c}{Camera Pose} & \multicolumn{3}{c}{Hand Pose}\\
    \cmidrule(lr){2-7}\cmidrule(lr){8-9}\cmidrule(lr){10-12}
    Method & PSNR$\uparrow$ & SSIM$\uparrow$ & LPIPS$\downarrow$ & FVD$\downarrow$ & Sub.Con.$\uparrow$ & BG.Con.$\uparrow$ & RotErr$\downarrow$ & TransErr$\downarrow$ & L2Err$\downarrow$ & PA-JPE$\downarrow$ & Recall$\uparrow$ \\
    \midrule
    Wan2.2-I2V-14B            & 14.97 & 0.4469 & 0.4609 & 479.12 & - & - & 11.14 & 8.11 & 83.47 & 9.50 & 88.82 \\
    ARCTIC-only               & 13.84 & 0.4453 & 0.4403 & 654.52 & 0.8870 & 0.8817 & 4.66 & 5.32 & 11.79 & 6.73 & 94.59 \\
    ARCTIC + EgoVid (raw)     & 14.87 & 0.4581 & 0.3997 & 550.20 & 0.9033 & 0.8938 & 4.03 & 4.91 & 13.08 & 7.56 & 91.85 \\
    \textbf{Ours (EgoVid-Pro)} & \textbf{17.42} & \textbf{0.5367} & \textbf{0.2923} & \textbf{274.51} & \textbf{0.9299} & \textbf{0.9229} & \textbf{3.75} & \textbf{3.33} & \textbf{11.32} & \textbf{6.37} & \textbf{94.75} \\
    \bottomrule
  \end{tabular}%
\end{table*}

\section{Experiments}
\label{sec:experiments}

\subsection{Experimental Setup}

\paragraph{Datasets.}
We propose \textbf{EgoVid-Pro}, a large-scale annotated egocentric dataset built upon EgoVid-5M~\citep{wang2024egovid5m}.
After filtering and annotation, the dataset comprises 103,032 video clips of 120 frames each with protagonist-centered 3D hand annotations, totaling approximately 12M annotated frames, comparable in scale to the largest existing egocentric dataset with 3D hand pose annotations~\citep{grauman2024egoexo4d} while spanning substantially more diverse everyday scenes.
The dataset captures diverse everyday activities with annotations covering both single-hand and bimanual interactions.
We extract a clean subset of 34,078 videos with complete bimanual annotations over the first 81 frames for training, reserving 300 videos as a held-out validation set.

To assess generalization beyond lab-controlled environments, we additionally evaluate on ARCTIC~\citep{fan2023arctic}, a multi-view motion capture dataset.
We adopt the standard train/test split of 267/34 videos (300--500 frames each) and center-crop the original 840$\times$600 resolution to 800$\times$600 to match our dataset's aspect ratio.
At each training step, we randomly sample an 81-frame clip per video;
while for evaluation, we extract three non-overlapping 81-frame clips from the beginning, middle, and end of each video.

\paragraph{Implementation Details.}
We initialize from the pretrained Wan2.2-I2V-14B checkpoint~\citep{wang2025wan} and apply parameter-efficient fine-tuning via LoRA~\citep{hu2021lora}.
Training proceeds for 1000 iterations using the AdamW optimizer with learning rate 1e-4 and batch size 16.
The control encoder comprises a 4-layer convolutional network that projects the $H \times W \times 12$ Pl\"{u}cker ray maps into the model's latent dimension.
All models are trained and evaluated at $480 \times 640$ resolution.
For reproducibility, all reported results are generated with a fixed random seed of $0$.

\paragraph{Evaluation Metrics.}
We assess model performance along three dimensions:
For \textbf{visual quality}, we report PSNR, SSIM, LPIPS~\citep{zhang2018lpips},
and Fr\'echet Video Distance (FVD), together with the subject-consistency
and background-consistency scores from VBench~\citep{huang2024vbench}.
For \textbf{hand-pose} and \textbf{camera-pose accuracy}, we reconstruct
the hand motion and camera trajectory from each generated video using
HaWoR~\citep{zhang2025hawor}, and evaluate the standard 2D and 3D hand
metrics (L2Err, PA-JPE) with the detection Recall, together with the
common camera metrics (RotErr, TransErr)~\citep{xie2026genreality}.
Following HaWoR~\citep{zhang2025hawor}, we further report a set of
world-space metrics (W-JPE, WA-JPE, RTE) that measure the absolute 3D
hand trajectory.

\begin{figure}[t]
  \centering
  \includegraphics[width=\linewidth]{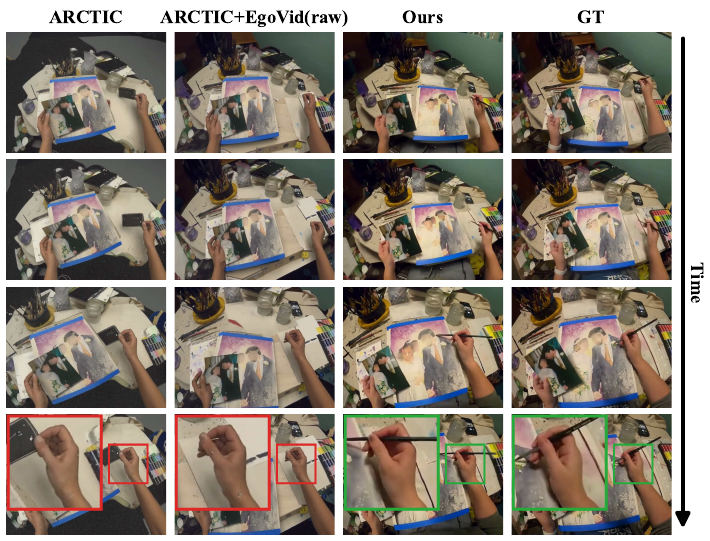}
  \caption{\textbf{Qualitative comparison of training-data conditions.}
    The ARCTIC-trained baselines drift toward lab imagery and place
    motion-capture markers on the synthesized hand (final-frame inset).
    Ours maintains the original scene appearance
    and produces more realistic hand interactions.}
  \label{fig:compare-data}
\end{figure}

\begin{table*}[t]
  \centering
  \caption{\textbf{Quantitative comparison with recent SOTA methods.}
    We achieve competitive performance against the baselines on ARCTIC. EgoVid-Pro brings out the full advantage of our representation, where we surpass all baselines on every metric. 
    \textbf{Bold} marks the best per column and
    \underline{underline} marks the second-best.
    }
  \label{tab:control_comparison}
  \tabcolsep=3pt
  \small
  \begin{tabular}{llccccccccccc}
    \toprule
    & & \multicolumn{4}{c}{Visual Quality} & \multicolumn{2}{c}{Camera Pose} & \multicolumn{5}{c}{Hand Pose}\\
    \cmidrule(lr){3-6}\cmidrule(lr){7-8}\cmidrule(lr){9-13}
    Dataset & Method & PSNR$\uparrow$ & SSIM$\uparrow$ & LPIPS$\downarrow$ & FVD$\downarrow$ & RotErr$\downarrow$ & TransErr$\downarrow$ & L2Err$\downarrow$ & PA-JPE$\downarrow$ & W-JPE$\downarrow$ & WA-JPE$\downarrow$ & RTE$\downarrow$ \\
    \midrule
    \multirow{5}{*}{ARCTIC}
      & Hand2World        & \textbf{16.93} & \textbf{0.6772} & \textbf{0.3309} & \underline{173.44} & \textbf{2.96} & \textbf{3.90} & \underline{15.20} & \underline{9.72} & \underline{118.45} & \underline{47.19} & \underline{8.63} \\
      & Generated Reality & 15.99 & 0.6627 & 0.3708 & \textbf{166.97} & \underline{3.11} & 4.24 & 15.39 & 9.93 & 132.07 & 47.37 & 12.12 \\
      & FMC*               & 16.51 & 0.6438 & 0.3842 & 403.13 & 3.60 & 4.36 & 23.48 & 10.98 & 148.44 & 57.02 & 12.25 \\
      & JointControl*       & 12.42 & 0.5852 & 0.5066 & 283.50 & 10.79 & 6.57 & 18.75 & 11.11 & 227.49 & 58.82 & 18.63 \\
      & \textbf{Ours}     & \underline{16.74} & \underline{0.6716} & \underline{0.3467} & 174.23 & 3.20 & \underline{4.14} & \textbf{14.36} & \textbf{9.34} & \textbf{114.00} & \textbf{46.91} & \textbf{7.68} \\
    \midrule
    \multirow{5}{*}{EgoVid-Pro}
      & Hand2World        & 16.68 & 0.5011 & 0.3438 & 336.33 & 5.56 & 5.59 & 21.84 & 8.50 & 76.06 & 41.66 & 5.44 \\
      & Generated Reality & 16.59 & 0.5069 & 0.3465 & 305.24 & 5.30 & 5.06 & 16.60 & 7.07 & \underline{75.82} & 37.08 & 5.07 \\
      & FMC*               & \underline{16.85} & \underline{0.5124} & \underline{0.3278} & 303.98 & \underline{4.76} & \underline{4.37} & \underline{13.51} & \underline{6.96} & 76.02 & \underline{36.66} & 5.34 \\
      & JointControl*       & 16.60 & 0.5051 & 0.3565 & \underline{298.56} & 7.44 & 6.08 & 17.08 & 7.33 & 79.05 & 40.95 & \underline{4.41} \\
      & \textbf{Ours}     & \textbf{17.42} & \textbf{0.5367} & \textbf{0.2923} & \textbf{274.51} & \textbf{3.75} & \textbf{3.33} & \textbf{11.32} & \textbf{6.37} & \textbf{67.46} & \textbf{33.02} & \textbf{4.12} \\
    \bottomrule
  \end{tabular}%
\end{table*}

\subsection{Effectiveness of EgoVid-Pro Annotations}
\label{sec:exp_data}

To demonstrate the effectiveness of our EgoVid-Pro annotations, we compare four training-data conditions:
\begin{itemize}[leftmargin=*,nosep]
  \item \textbf{Wan2.2-I2V-14B}: the pretrained zero-control reference.
  \item \textbf{ARCTIC-only}: fine-tuned on the ARCTIC dataset alone.
  \item \textbf{ARCTIC + EgoVid (raw)}: pretrained on unannotated EgoVid
    clips and then fine-tuned on ARCTIC.
  \item \textbf{Ours (EgoVid-Pro)}: fine-tuned on our
    protagonist-annotated EgoVid-Pro dataset.
\end{itemize}

\Cref{tab:data_effectiveness} reports the quantitative comparison.
Pretraining on raw EgoVid clips improves visual quality over ARCTIC
alone, but it also hurts hand-control accuracy: the unfiltered video
adds scene diversity at the cost of label quality.
By extending clean, large-scale annotations to in-the-wild scenes,
our EgoVid-Pro avoids this tradeoff and achieves the best on
both visual quality and control accuracy.

\Cref{fig:compare-data} visualizes the same hand-trajectory condition
rolled out by each model on an unseen validation clip.
The ARCTIC-based variants fail in two aspects:
\textbf{(i) Distribution leakage.}
Despite first-frame conditioning, ARCTIC-only and
ARCTIC+EgoVid~(raw) drift toward the ARCTIC distribution within a few
seconds: the color tone shifts to lab lighting, and both the background
and the on-table objects start to look like ARCTIC content.
The hand inset on the final frame is the clearest case.
The generated hand carries the mocap markers on the skin, which
never appear in the input clip.
\textbf{(ii) Object confusion.}
In this example, the protagonist's left hand holds a photograph next to
a drawing board on the table. Both ARCTIC baselines treat the
photograph and the board as a single rigid object and translate them
together, while ours moves only the photograph and leaves the board in
place.
More examples will be shown in the supplement material. 

\subsection{Comparison of Control Signals}
\label{sec:exp_control}

\begin{figure}[t]
  \centering
  \includegraphics[width=\linewidth]{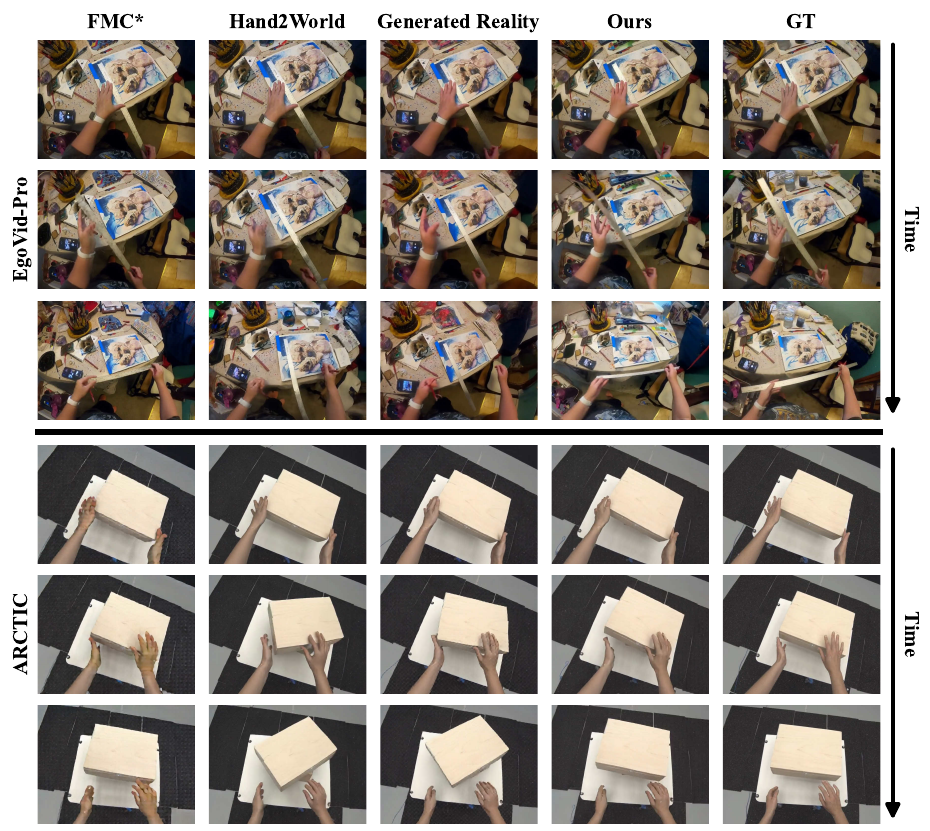}
  \caption{\textbf{Qualitative comparison of control signal representations}
    on EgoVid-Pro (picking up a ruler) and ARCTIC (using a box).
    The camera-space baselines misplace
    the hand and miss the intended contact. Our Pl\"{u}cker-ray
    representation produces the most realistic hand interactions.}
  \label{fig:compare-method}
\end{figure}

\begin{table*}[t]
  \centering
  \caption{Comparison of protagonist-filtering strategies and EgoSim on
    EgoVid-Pro. \textbf{Bold} and \underline{underline} mark the best and
    second-best values per column, respectively.}
  \label{tab:data_filtering_ablation}
  \tabcolsep=2.5pt
  \small
  \begin{tabular}{lccccccccccc}
    \toprule
    & \multicolumn{6}{c}{Visual Quality} & \multicolumn{2}{c}{Camera Pose} & \multicolumn{3}{c}{Hand Pose}\\
    \cmidrule(lr){2-7}\cmidrule(lr){8-9}\cmidrule(lr){10-12}
    Strategy & PSNR$\uparrow$ & SSIM$\uparrow$ & LPIPS$\downarrow$ & FVD$\downarrow$ & Sub.Con.$\uparrow$ & BG.Con.$\uparrow$ & RotErr$\downarrow$ & TransErr$\downarrow$ & L2Err$\downarrow$ & PA-JPE$\downarrow$ & Recall$\uparrow$ \\
    \midrule
    Camera-outward heuristic & \underline{16.46} & \underline{0.4981} & \underline{0.3547} & \underline{327.36} & 0.9293 & 0.9155 & 6.52 & 5.66 & \underline{15.49} & \underline{7.39} & \underline{93.26} \\
    MCP-up heuristic         & 14.28 & 0.4262 & 0.4916 & 542.65 & \textbf{0.9309} & 0.9188 & 11.41 & 8.44 & 93.92 & 11.05 & 83.74 \\
    EgoSim~\citep{hao2026egosim} & 13.95 & 0.3987 & 0.4492 & 660.17 & \underline{0.9306} & \textbf{0.9313} & \textbf{2.99} & \textbf{3.17} & 44.26 & 11.41 & 75.83 \\
    \textbf{Ours}            & \textbf{17.42} & \textbf{0.5367} & \textbf{0.2923} & \textbf{274.51} & 0.9299 & \underline{0.9229} & \underline{3.75} & \underline{3.33} & \textbf{11.32} & \textbf{6.37} & \textbf{94.75} \\
    \bottomrule
  \end{tabular}%
\end{table*}

To isolate the contribution of our Pl\"{u}cker-ray representation, we conduct controlled comparisons against recent hand-controlled generation methods on ARCTIC dataset and our proposed EgoVid-Pro dataset, respectively.
For Hand2World~\citep{wang2026hand2world} and Generated Reality~\citep{xie2026genreality}, which also focus on egocentric video generation, we faithfully reimplement their approaches following the original papers and open-source codes.
For FMC~\citep{shuai2025fmc}, which encodes object 6D poses within object silhouettes originally, we adapt their approach to hands by encoding 16 hand joints with 9D pose descriptors (FMC*).
Since JointControl~\citep{zhang2026occlusionhand} focuses on static-camera settings only, we reimplement their approach augmented with the same camera control branch as ours for a fair comparison (JointControl*).

\Cref{tab:control_comparison} reports the quantitative comparison.
We match the strongest baselines on ARCTIC, while surpassing all of
them on every metric on the more diverse EgoVid-Pro.
The contrast stems from the limited ego-motion of tabletop captures:
with the camera frame nearly aligned with the world frame,
camera-space and world-space encodings carry essentially the same
signal, and no choice of representation yields a measurable advantage.
Under substantial ego-motion, however, the two diverge and the
world-space formulation becomes essential.
EgoVid-Pro provides precisely this regime, underscoring its diagnostic
value as a benchmark.

We further illustrate this contrast in \Cref{fig:compare-method} with one representative case from each dataset.
The camera-space encodings (Hand2World, Generated Reality) entangle the
camera trajectory with the hand articulation, and the rendered hand
position drifts from the conditioning trajectory.
In the EgoVid-Pro case, the generated hand misses the ruler on the
table; in the ARCTIC case, it appears to grasp a box without actually
touching it.
As a reimplemented world-space baseline, FMC* recovers the correct contact
in both cases,
which shows that the world-space property, rather than the specific
encoding, removes this entanglement.
Our Pl\"{u}cker Hand Map realizes this property with a dense per-pixel
signal rather than a compressed per-joint descriptor, yielding the
most accurate hand motion among all methods.

\subsection{Ablation Study}
\label{sec:exp_data_filtering}

To isolate the effectiveness of our protagonist-centered data filtering strategy, we compare with the camera-outward and
MCP-up heuristics discussed in the main text, as well as
EgoSim~\citep{hao2026egosim}. EgoSim similarly annotates large-scale
egocentric data to improve generalization, but restricts its data to tabletop
scenes and therefore does not perform any filtering for unconstrained scenes. Moreover, it reconstructs the scene from the first frame
and reprojects it along the target camera trajectory as a dense control signal,
providing a comparatively static scene representation.

\begin{figure}[t]
  \centering
  \includegraphics[width=\linewidth]{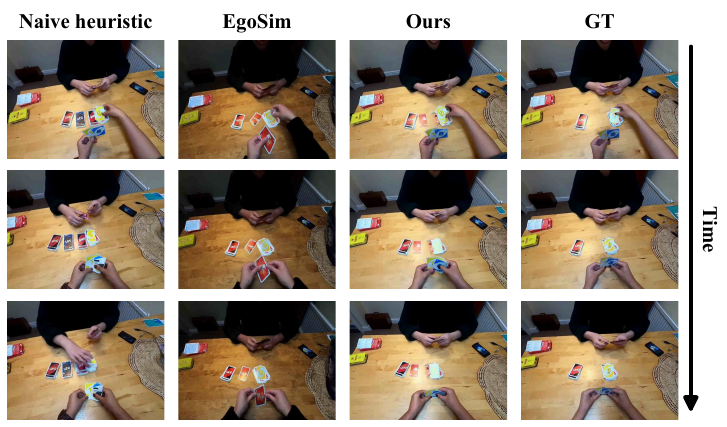}
  \caption{\textbf{Ablation study of protagonist-centered data
    filtering.} Naive filtering introduces ambiguity between the controlled
    protagonist and the other player, while EgoSim's static scene
    representation suppresses the other player's natural motion. Ours follows
    the protagonist's hand trajectory while preserving natural environment
    dynamics.}
  \label{fig:data_filtering_ablation}
\end{figure}

As shown in \cref{tab:data_filtering_ablation}, our approach yields the
strongest overall visual quality and hand-control accuracy. More importantly,
protagonist-centered annotation establishes a clear control boundary: the
protagonist's hands should follow the geometric trajectory, while the rest of
the environment retains its natural dynamics. Static objects should remain
stable, whereas other people and animals may move naturally or respond to the
text prompt.
Bystander trajectories would contaminate the training signal and blur this boundary.
In the UNO-card case in 
\cref{fig:data_filtering_ablation}, the prompt describes the protagonist
shuffling cards while the opposing player holds a hand of cards. After the
protagonist finishes shuffling and keeps both hands still, the naive
heuristic incorrectly transfers the text-described shuffling motion to the
opposing player. EgoSim exhibits a different failure mode: its static scene
representation tightly constrains camera geometry but suppresses natural
environment dynamics. The opposing player consequently appears frozen,
whereas our model preserves subtle natural motion, such as breathing and
adjusting the held cards, while keeping the protagonist's action under
geometric control.

\section{Conclusion}

We presented \method, bringing hand-controlled egocentric video generation from lab settings to unconstrained everyday scenes.
Our annotation pipeline distills clean protagonist-only trajectories from in-the-wild video via 3-level filtering, yielding EgoVid-Pro.
Our Plücker Hand Map encodes camera and hands as world-space rays, disentangling hand motion from ego-motion and keeping the absolute 3D hand trajectory unambiguous.
These contribute to training on diverse monocular video and generalization to everyday egocentric interaction.



{
\small
\bibliographystyle{ieeenat_fullname}
\bibliography{refs}
}

\clearpage
\maketitlesupplementary
\appendix
\setcounter{table}{0}
\setcounter{figure}{0}
\setcounter{equation}{0}
\renewcommand{\thetable}{\Alph{table}}
\renewcommand{\thefigure}{\Alph{figure}}
\renewcommand{\theequation}{\Alph{equation}}
\renewcommand{\theHtable}{supp.table.\arabic{table}}
\renewcommand{\theHfigure}{supp.figure.\arabic{figure}}
\renewcommand{\theHequation}{supp.equation.\arabic{equation}}


\section{Additional In-the-Wild Results}
\label{sec:supp_showcases}

The main paper presents a gallery of \emph{fully in-the-wild} generation
results (\cref{fig:showcases}): neither the first frames nor the conditioning
hand trajectories come from a captured test set.
These examples cover not only ordinary everyday environments but also
imaginary and stylized worlds with no counterpart in any egocentric capture
corpus.
They also span a wide range of actions, from reaching and grasping to lifting,
pushing, and bimanual coordination, all under substantial camera ego-motion and
all produced by a single set of model weights.
Beyond placing the hands at the prescribed location, the model faithfully
synthesizes their effect on the scene, where contact, occlusion, and object
motion are consistent with the conditioning hand geometry.

Such open-ended control follows directly from our two contributions.
Conditioning on world-space Pl\"{u}cker rays specifies the hand in a metric 3D
frame disentangled from ego-motion, so a given trajectory denotes the same
physical hand motion regardless of how the camera moves; camera-space signals
instead entangle the two motions in the image plane, leaving the intended 3D
motion ambiguous.
EgoVid-Pro contributes the other half.
It provides clean and protagonist-only 3D hand annotations from
large-scale unconstrained video rather than from a fixed tabletop capture.
Training therefore exposes the model to a far broader distribution of scenes,
objects, and interactions, enabling the same control transfer to environments
never seen during training.
Prior hand-controlled generators entangled camera-space conditioning with
tabletop training data, and thus possess neither property.

\begin{figure*}[t]
  \centering
  \includegraphics[width=\linewidth]{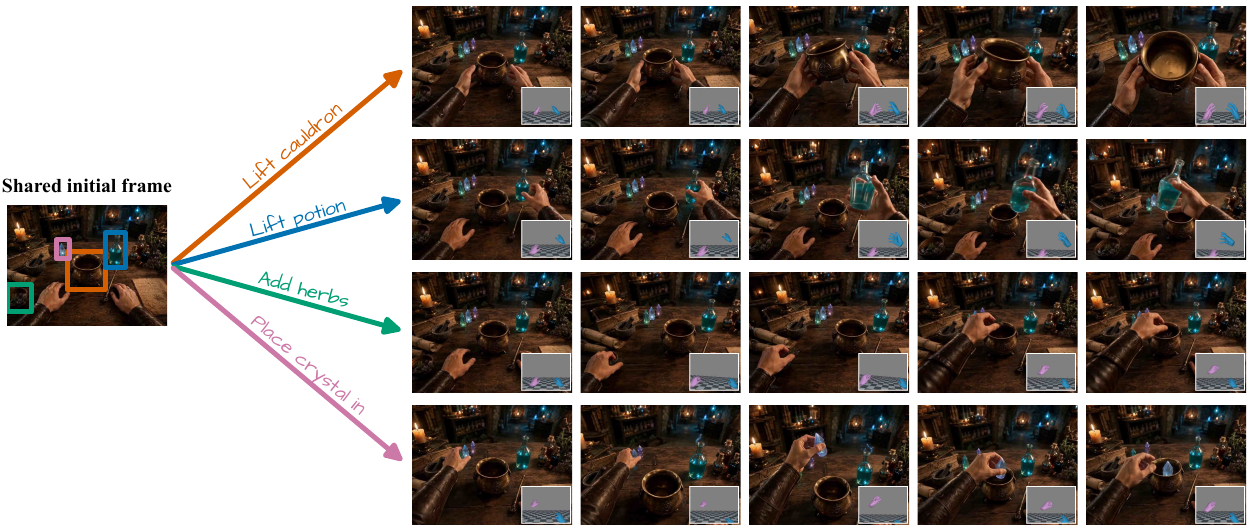}
  \caption{\textbf{Diverse interactions from a shared initial scene.}
    From the same first frame, our method generates four distinct interactions
    targeting different objects. The pose insets visualize the corresponding
    hand controls.}
  \label{fig:same_scene_actions}
\end{figure*}

Beyond generalizing across scenes, our method can synthesize diverse
interactions within the same environment. As shown in
\cref{fig:same_scene_actions}, starting from a shared initial frame, we apply
four distinct action controls to different objects in the scene. The generated
videos faithfully realize the specified motions and produce coherent
interactions with the corresponding objects, demonstrating that the model does
not merely reproduce a single scene-specific motion pattern.

\section{Details of SMPL Body Fitting}
\label{sec:supp_smpl_fit}

This section provides the complete formulation of the world-space body-fit
optimization summarized in \cref{sec:data_3d}.

\paragraph{Inputs.}
HaWoR predicts each hand tracklet in camera-space coordinates, \ie,
relative to a camera with identity extrinsics. For each frame $t$ along
the tracklet, this yields two camera-space hand anchors $\mathbf{Y}_t =
\{\mathbf{y}^{w}_t, \mathbf{y}^{m}_t\}$: the wrist position
$\mathbf{y}^{w}_t$ and a middle-finger MCP-direction anchor
$\mathbf{y}^{m}_t$.

\paragraph{Variables.}
All tracklets in a clip are fit jointly with a single shared body shape
$\boldsymbol{\beta}$. For each frame $t$ we optimize a
VPoser~\citep{pavlakos2019expressive} latent body code $\mathbf{z}_t$, a
global orientation $\mathbf{R}_t$, and a global translation
$\mathbf{t}_t$. The articulated pose is decoded as
$\boldsymbol{\theta}_t = D_{\mathrm{VPoser}}(\mathbf{z}_t)$ and combined
with $\mathbf{R}_t,\mathbf{t}_t,\boldsymbol{\beta}$ through the SMPL
forward map to produce the joint and vertex sets
\begin{equation}
(\mathbf{J}_t, \mathbf{V}_t)
= M(\boldsymbol{\theta}_t, \mathbf{R}_t, \mathbf{t}_t, \boldsymbol{\beta}).
\end{equation}

\paragraph{Head loss.}
The head loss enforces that the fitted head pose coincides with the
identity camera, placing the head at the origin with the gaze along $+z$
and the inter-eye axis along $+x$. Let $\mathbf{V}^{L\text{-eye}}_t,
\mathbf{V}^{R\text{-eye}}_t$ denote two manually pre-selected vertices on
the SMPL face mesh corresponding to the left and right eye centers. The
head constraint has three components: the eye center
$\mathbf{e}_t = \tfrac{1}{2}(\mathbf{V}^{L\text{-eye}}_t + \mathbf{V}^{R\text{-eye}}_t)$
is anchored at the origin; the body gaze direction $\mathbf{g}_t$,
defined as the averaged outward face normal over a small manually
selected set of forward-facing eye-region vertices, is aligned with
$\mathbf{g}^{*}=[0,0,1]$; and the inter-eye direction
$\mathbf{r}_t = (\mathbf{V}^{R\text{-eye}}_t - \mathbf{V}^{L\text{-eye}}_t) /
\|\mathbf{V}^{R\text{-eye}}_t - \mathbf{V}^{L\text{-eye}}_t\|$ is aligned
with $\mathbf{r}^{*}=[1,0,0]$:
\begin{equation}
\mathcal{L}_{\mathrm{head}}
= \sum_t \|\mathbf{e}_t\|_2^2
+ \tfrac{1}{T}\sum_t (1 - \mathbf{g}_t^\top \mathbf{g}^{*})
+ \tfrac{1}{T}\sum_t (1 - \mathbf{r}_t^\top \mathbf{r}^{*}),
\end{equation}
where $T$ is the number of frames in the clip's joint optimization.

\paragraph{Hand loss.}
The hand loss aligns the SMPL wrist and middle-finger-MCP joints
$\mathbf{j}^{w}_t, \mathbf{j}^{m}_t$ (selected by handedness) with the
corresponding HaWoR anchors:
\begin{equation}
\mathcal{L}_{\mathrm{hand}}
= \sum_t \bigl( \|\mathbf{j}^{w}_t - \mathbf{y}^{w}_t\|_2^2
              + \|\mathbf{j}^{m}_t - \mathbf{y}^{m}_t\|_2^2 \bigr).
\end{equation}

\paragraph{Regularization and full objective.}
We further regularize the VPoser latent and the shared shape parameters
with $\mathcal{L}_{\mathrm{pose}} = \sum_t \|\mathbf{z}_t\|_2^2$ and
$\mathcal{L}_{\mathrm{shape}} = \|\boldsymbol{\beta}\|_2^2$. The full
objective
\begin{equation}
\mathcal{L}
= \lambda_{\mathrm{data}}
\bigl( \mathcal{L}_{\mathrm{head}} + \mathcal{L}_{\mathrm{hand}} \bigr)
+ \lambda_{\mathrm{pose}} \mathcal{L}_{\mathrm{pose}}
+ \lambda_{\mathrm{shape}} \mathcal{L}_{\mathrm{shape}}
\label{eq:smpl_opt}
\end{equation}
is jointly minimized over all per-frame variables with Adam, holding the
SMPL model and the VPoser decoder fixed. We set
$\lambda_{\mathrm{data}} = 10$,
$\lambda_{\mathrm{pose}} = 5\!\times\!10^{-3}$, and
$\lambda_{\mathrm{shape}} = 3\!\times\!10^{-2}$.

\paragraph{Gap-filling threshold.}
Before the linear-interpolation step that fills frames lacking valid detections (\cref{sec:data_3d}), we verify that the two bracketing
detections describe similar hand poses by requiring both their mean
per-joint L2 distance and their mean per-vertex L2 distance to fall below
$\tau_{\mathrm{gap}} = 0.4$\,m. Clips that fail this check are discarded
entirely.

\begin{table*}[t]
  \centering
  \caption{\textbf{Out-of-distribution quantitative comparison on H2O.}
    All methods are trained on the same EgoVid-Pro data and evaluated directly
    on H2O. FMC* is included in the quantitative evaluation but omitted from
    the qualitative comparison. \textbf{Bold} marks the best result and
    \underline{underline} the second-best.}
  \label{tab:h2o_ood}
  \tabcolsep=3pt
  \small
  \begin{tabular}{lccccccccccc}
    \toprule
    & \multicolumn{4}{c}{Visual Quality} & \multicolumn{2}{c}{Camera Pose} & \multicolumn{5}{c}{Hand Pose}\\
    \cmidrule(lr){2-5}\cmidrule(lr){6-7}\cmidrule(lr){8-12}
    Method & PSNR$\uparrow$ & SSIM$\uparrow$ & LPIPS$\downarrow$ & FVD$\downarrow$ & RotErr$\downarrow$ & TransErr$\downarrow$ & L2Err$\downarrow$ & PA-JPE$\downarrow$ & W-JPE$\downarrow$ & WA-JPE$\downarrow$ & RTE$\downarrow$ \\
    \midrule
    Hand2World        & 15.35 & 0.4915 & 0.4369 & 557.61 & 3.44 & 1.94 & 41.39 & 9.64 & 139.16 & 41.45 & 8.34 \\
    Generated Reality & 15.49 & \underline{0.5141} & 0.4176 & 557.28 & 3.60 & 1.90 & \underline{24.12} & 11.46 & \underline{84.22} & 44.34 & \underline{4.87} \\
    FMC*              & \underline{15.65} & 0.5048 & \underline{0.4141} & 485.71 & \underline{2.87} & \underline{1.80} & 40.95 & \underline{9.09} & 137.25 & 40.06 & 7.25 \\
    JointControl*     & 15.51 & 0.5005 & 0.4427 & \underline{442.74} & 4.55 & 2.22 & 40.44 & 9.82 & 104.55 & \underline{40.04} & 6.46 \\
    \textbf{Ours}    & \textbf{17.98} & \textbf{0.5902} & \textbf{0.2270} & \textbf{343.86} & \textbf{1.05} & \textbf{0.93} & \textbf{9.98} & \textbf{7.83} & \textbf{59.51} & \textbf{30.04} & \textbf{4.58} \\
    \bottomrule
  \end{tabular}%
\end{table*}

\begin{figure*}[t]
  \centering
  \includegraphics[width=0.93\linewidth]{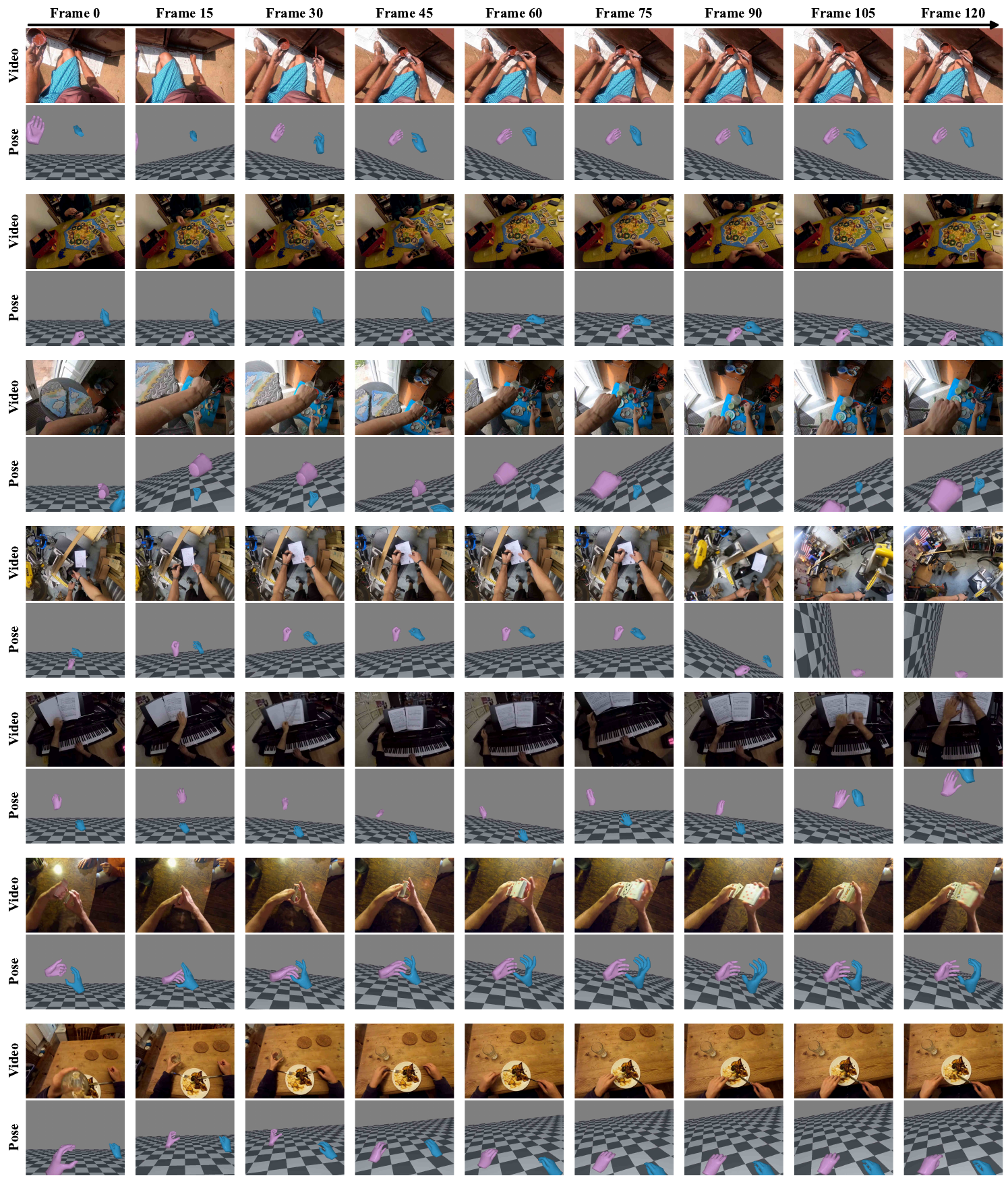}
  \caption{\textbf{Annotation results on EgoVid-Pro.} Each case shows a source
    clip (\emph{Video}) uniformly sampled at nine timesteps, with our recovered
    world-space hand pose (\emph{Pose}) rendered beneath the corresponding
    frame. The ground checkerboard, anchored to the first-frame world frame,
    visualizes the recovered camera trajectory.}
  \label{fig:supp_annotation}
\end{figure*}

\paragraph{Annotation showcases.}
To illustrate the output of the annotation pipeline described above, we
showcase several EgoVid-Pro clips together with the annotated camera and hand poses.
\Cref{fig:supp_annotation} presents a gallery of representative examples. For
each clip, we uniformly sample nine frames spanning the sequence and, beneath
each source frame, render the corresponding fitted hand mesh within the
world-space coordinate frame used for conditioning. To make the estimated
camera pose visually interpretable, we further overlay a checkerboard ground
plane in each pose rendering. Since our calibration procedure does not
estimate the absolute position of the physical ground, we place this reference
plane $1.25$\,m below the first-frame camera; it spans $18\,\mathrm{m}\times
18\,\mathrm{m}$ with a tile size of $0.5\,\mathrm{m}\times 0.5\,\mathrm{m}$.
Across diverse scenes, viewpoints, and hand--object interactions, the
recovered poses remain temporally stable and stay tightly aligned with the
protagonist's hands throughout each clip, confirming the quality and
consistency of the annotations that drive our geometric control signal.

\begin{figure}[t]
  \centering
  \includegraphics[width=0.9\linewidth]{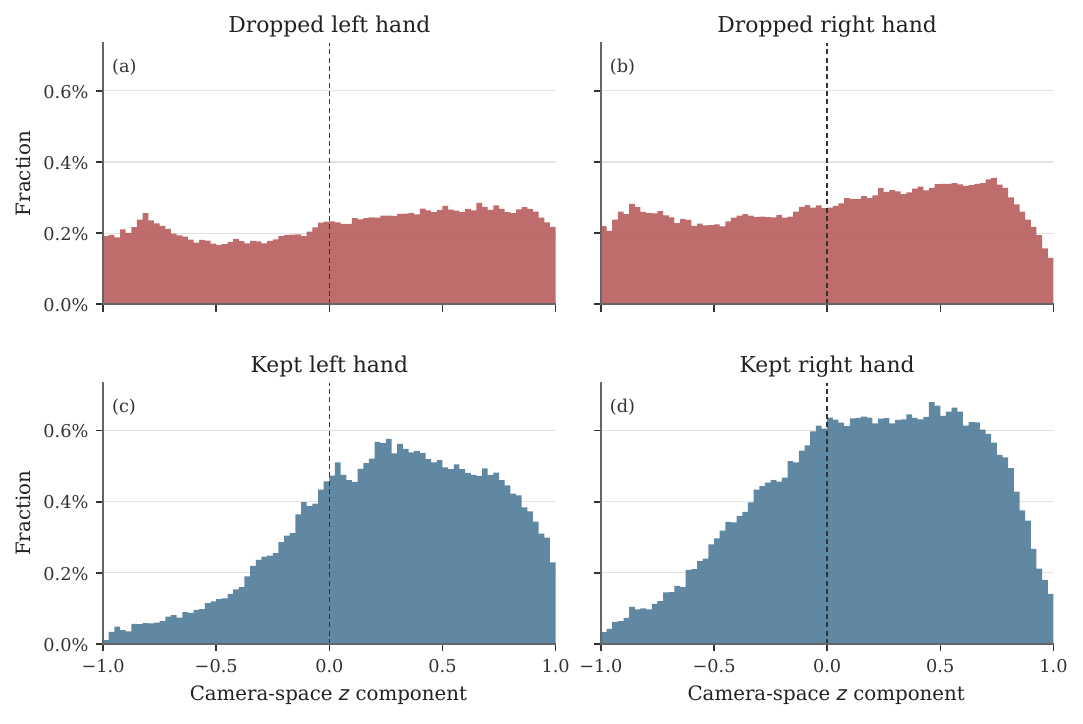}
  \caption{\textbf{Camera-space hand-orientation distributions.}
    Histograms show the $z$ component of the hand orientation for left and
    right tracklets rejected (top) or retained (bottom) by our geometry-based
    filter. Retained hands follow the expected egocentric forward-facing bias,
    yet rejected tracklets overlap substantially with the same range,
    demonstrating that a naive camera-space criterion cannot cleanly identify
    the protagonist's hands.}
  \label{fig:supp_z_hist}
\end{figure}

\begin{figure}[t]
  \centering
  \includegraphics[width=\linewidth]{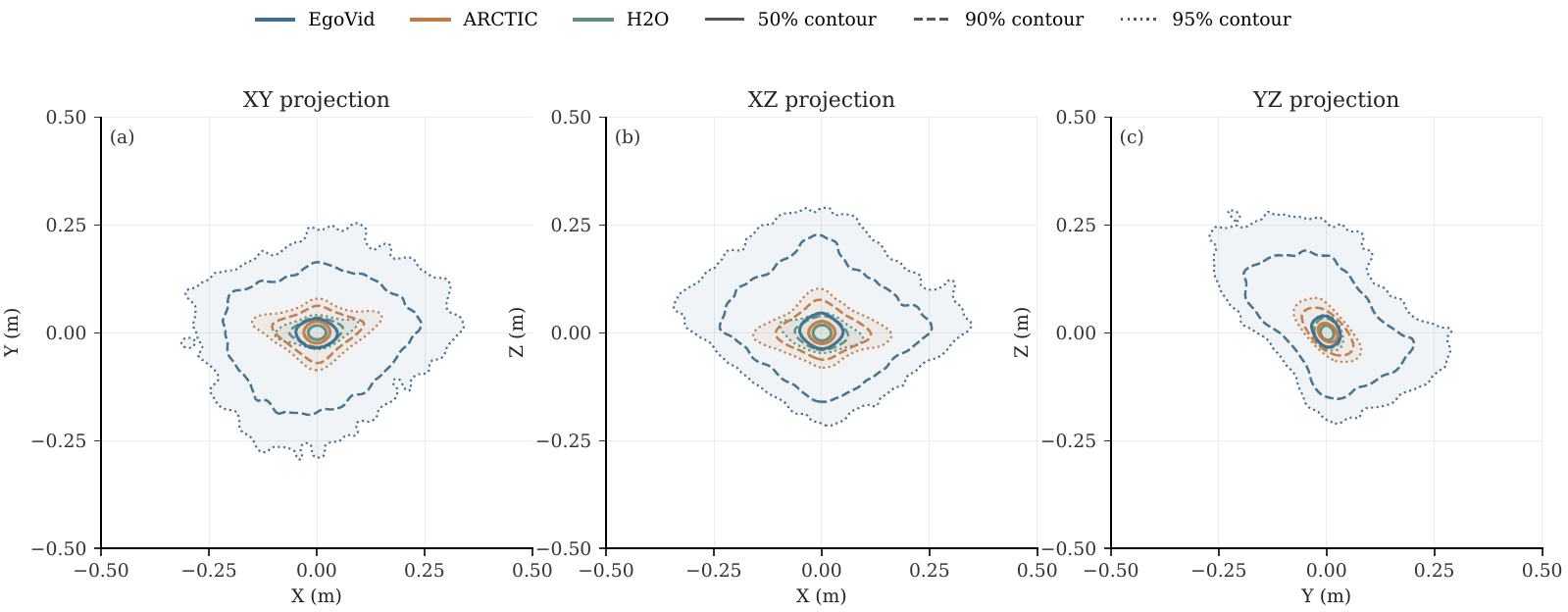}
  \caption{\textbf{Camera-position distributions across datasets.}
    The three panels show
    orthographic projections of the aggregated origins, using identical axis
    limits and tick locations over $[-0.5,0.5]$\,m; points outside this display
    range are clipped. Solid, dashed, and dotted curves denote the 50\%, 90\%,
    and 95\% highest-density regions, respectively. EgoVid-Pro covers a
    markedly broader range than the tabletop-centered
    ARCTIC~\citep{fan2023arctic} and H2O~\citep{kwon2021h2o} datasets.}
  \label{fig:supp_camera_distribution}
\end{figure}

\paragraph{Dataset statistics.}
We provide additional statistics that characterize both our filtering
problem and the motion diversity of EgoVid-Pro. As shown in
\cref{fig:supp_z_hist}, retained left and right hands exhibit the expected
egocentric tendency to point forward in camera space. However, a substantial
fraction of rejected tracklets occupies the same positive-$z$ region. The
large overlap between the retained and rejected distributions explains why a
naive threshold on camera-space orientation cannot reliably distinguish the
protagonist's hands from bystander hands and false detections.

We further compares camera motion across
datasets. To quantify camera-motion diversity, we transform the camera centers
of all 81 frames in each window into the coordinate system of its first frame,
and aggregate the resulting relative camera positions over all windows in each
dataset. \Cref{fig:supp_camera_distribution} visualizes their density in the
three orthographic planes. EgoVid-Pro spans a substantially broader range in
all three projections, whereas
ARCTIC~\citep{fan2023arctic} and H2O~\citep{kwon2021h2o} remain tightly
concentrated around the initial camera, reflecting their predominantly
tabletop capture settings. This diversity
exposes our model to considerably richer ego-motion during training and
supports generalization beyond fixed laboratory layouts.

\begin{figure*}[t]
  \centering
  \includegraphics[width=\linewidth]{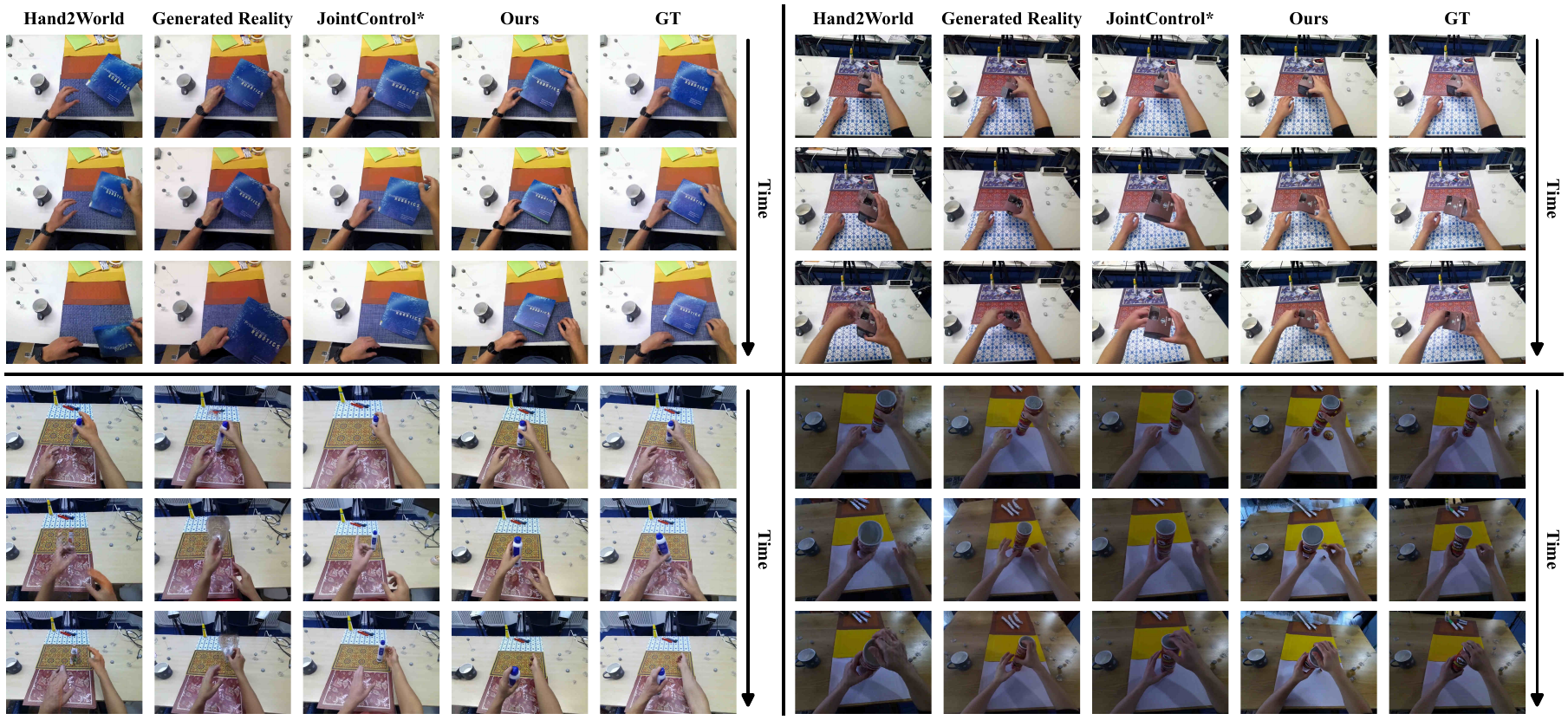}
  \caption{\textbf{Qualitative comparison on the out-of-distribution H2O
    dataset.} All methods use the same EgoVid-Pro training data. Our method  generates more
    accurate and realistic hand--object interactions.}
  \label{fig:h2o_ood}
\end{figure*}

\section{Training Details}
\label{sec:training}

\paragraph{CFG conditioning.}
We train with classifier-free guidance~\citep{ho2022cfg}.
The 12-channel Pl\"{u}cker map is treated as a single \emph{geometric}
conditioning signal: during training, we independently drop the text
condition with probability $p_{\text{text}} = 0.1$ and the geometric
condition with
probability $p_{\text{geo}} = 0.1$.
At inference, a single pair of guidance scales $w_{\text{text}},
w_{\text{geo}}$ trades off text fidelity against geometric accuracy.

\paragraph{Autoregressive distillation.}
For temporally consistent long-video inference beyond the model's native
context window, we follow a causal-forcing distillation
strategy~\citep{huang2025selfforcing, zhu2026causal}: a student model is trained to
match the distribution of the base diffusion model~\citep{yin2024dmd} when conditioned on frames
generated by the student itself.
This eliminates exposure-bias artifacts in autoregressive rollout without
requiring expensive multi-step diffusion inference at every window.
We apply this distillation to the lightweight 5B backbone, increasing
autoregressive generation speed to 9.64 FPS while maintaining competitive
visual quality.

\begin{figure*}[!t]
  \centering
  \includegraphics[width=\linewidth]{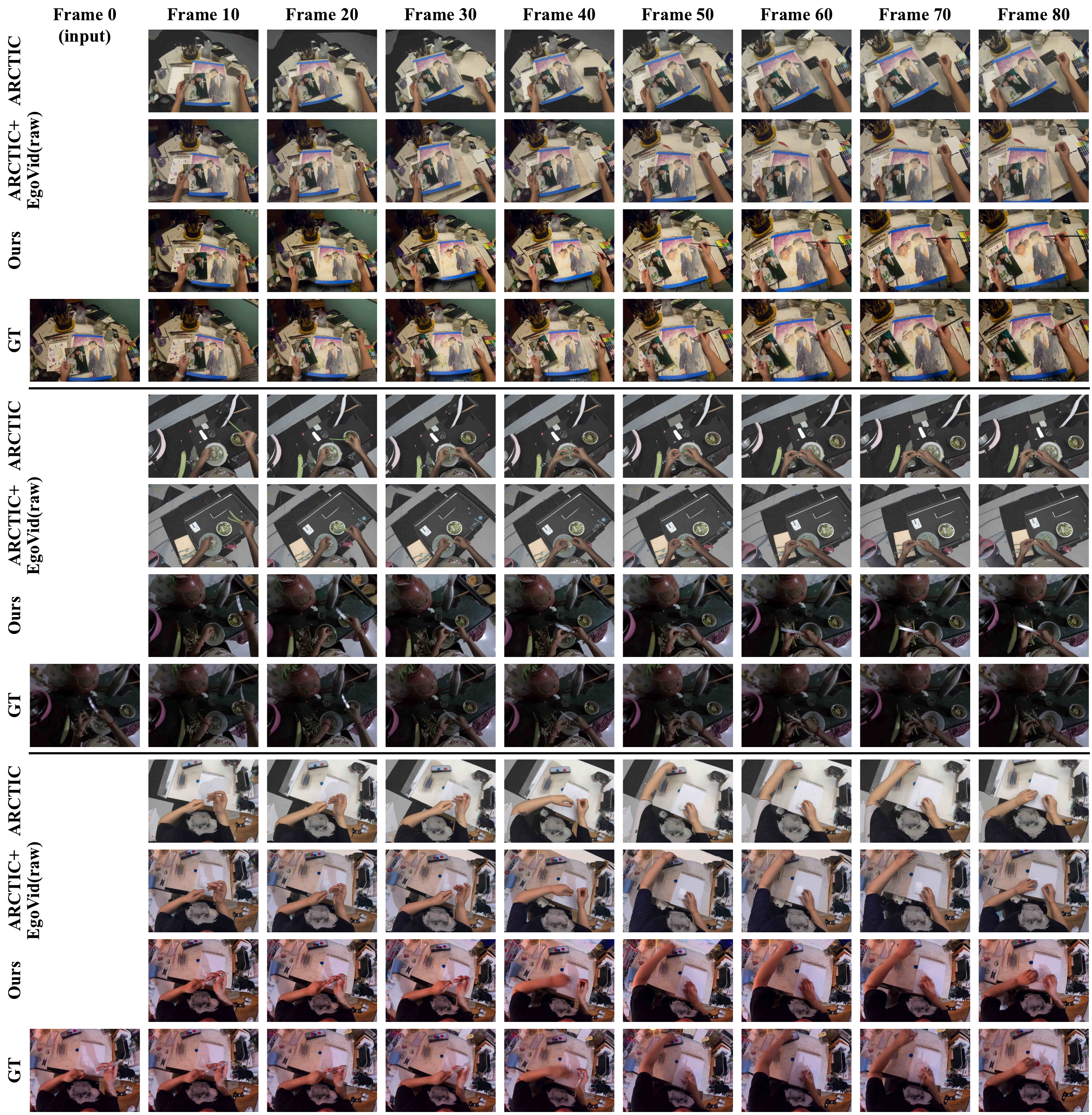}
  \caption{\textbf{Additional qualitative comparison of training-data
    conditions.} Extending \cref{fig:compare-data} with more frames and examples.}
  \label{fig:supp_more_data}
\end{figure*}

\begin{table}[t]
  \centering
  \caption{\textbf{Quality and inference speed of autoregressive
    distillation.} FPS is measured under the same 81-frame inference setting.
    \textbf{Bold} and \underline{underline} denote the best and second-best
    results, respectively.}
  \label{tab:ar_distillation}
  \setlength{\tabcolsep}{4.5pt}
  \small
  \begin{tabular}{lccccc}
    \toprule
    Model & PSNR$\uparrow$ & SSIM$\uparrow$ & LPIPS$\downarrow$ & FVD$\downarrow$ & FPS$\uparrow$ \\
    \midrule
    14B    & \textbf{17.42} & \textbf{0.5367} & \textbf{0.2923} & \textbf{274.51} & 0.13 \\
    5B     & \underline{16.69} & \underline{0.4962} & \underline{0.3663} & 384.01 & \underline{2.39} \\
    5B-AR  & 16.31 & 0.4829 & \underline{0.3663} & \underline{354.36} & \textbf{9.64} \\
    \bottomrule
  \end{tabular}
\end{table}

As shown in \cref{tab:ar_distillation}, the 5B backbone improves inference
efficiency from 0.13 to 2.39 FPS, and autoregressive distillation further increases
it to 9.64 FPS. Despite this substantial acceleration, 5B-AR maintains
comparable visual quality to the undistilled 5B model, demonstrating an
effective quality--efficiency trade-off for long-video generation.

\section{More Experiments}

\begin{table*}[t]
  \centering
  \caption{Comparison of different text CFG guidance scales.
    The geometric guidance scale is fixed at $w_{\mathrm{geo}}=2$.
    \textbf{Bold} marks the best per column; \underline{underline} marks
    the second-best (next distinct value when ties are bolded).}
  \label{tab:supp_cfg_text}
  \tabcolsep=4pt
  \small
  \begin{tabular}{cccccccccccc}
    \toprule
    & \multicolumn{6}{c}{Visual Quality} & \multicolumn{2}{c}{Camera Pose} & \multicolumn{3}{c}{Hand Pose}\\
    \cmidrule(lr){2-7}\cmidrule(lr){8-9}\cmidrule(lr){10-12}
    $w_{\mathrm{text}}$ & PSNR$\uparrow$ & SSIM$\uparrow$ & LPIPS$\downarrow$ & FVD$\downarrow$ & Sub.Con.$\uparrow$ & BG.Con.$\uparrow$ & RotErr$\downarrow$ & TransErr$\downarrow$ & L2Err$\downarrow$ & PA-JPE$\downarrow$ & Recall$\uparrow$ \\
    \midrule
    1 & 17.39 & 0.5354 & 0.2943 & \textbf{266.82} & \underline{0.9305} & 0.9224 & \textbf{3.73} & \underline{3.35} & \underline{11.83} & \underline{6.43} & 94.62 \\
    2 & \textbf{17.43} & \textbf{0.5367} & 0.2928 & \underline{274.45} & \textbf{0.9308} & \textbf{0.9240} & 3.84 & 3.49 & 16.10 & 6.47 & 94.17 \\
    3 & \underline{17.42} & \textbf{0.5367} & \underline{0.2923} & 274.51 & 0.9299 & 0.9229 & \underline{3.75} & \textbf{3.33} & \textbf{11.32} & \textbf{6.37} & \underline{94.75} \\
    4 & 17.41 & \underline{0.5363} & \textbf{0.2922} & 283.53 & 0.9283 & \underline{0.9231} & 4.00 & 3.61 & 11.92 & 6.47 & \textbf{94.76} \\
    5 & 17.39 & 0.5357 & 0.2924 & 282.80 & 0.9271 & 0.9226 & 3.82 & 3.48 & 12.60 & 6.49 & 94.21 \\
    \bottomrule
  \end{tabular}%
\end{table*}

\paragraph{Out-of-distribution evaluation.}
We further evaluate generalization in a fully out-of-distribution setting.
All methods are trained on the same EgoVid-Pro dataset and tested directly on
H2O~\citep{kwon2021h2o}, without seeing its tabletop layouts or objects during
training. 
Notably, H2O uses slightly different camera intrinsics, which introduce an
additional domain shift.
Moreover, unlike the in-domain evaluation on our annotated data, H2O provides independently obtained multi-view hand-pose annotations, enabling a fairer and less biased assessment of how faithfully the generated hands follow the control trajectories.
As shown in \cref{tab:h2o_ood,fig:h2o_ood}, our method generalizes robustly
under these substantial domain shifts. Despite being trained on exactly the
same large-scale dataset as the baselines, our method better represents the
spatial relationship between the camera and hands, leading to more faithful
control and more realistic hand--object interaction generation.

\begin{figure}[t]
  \centering
  \includegraphics[width=\linewidth]{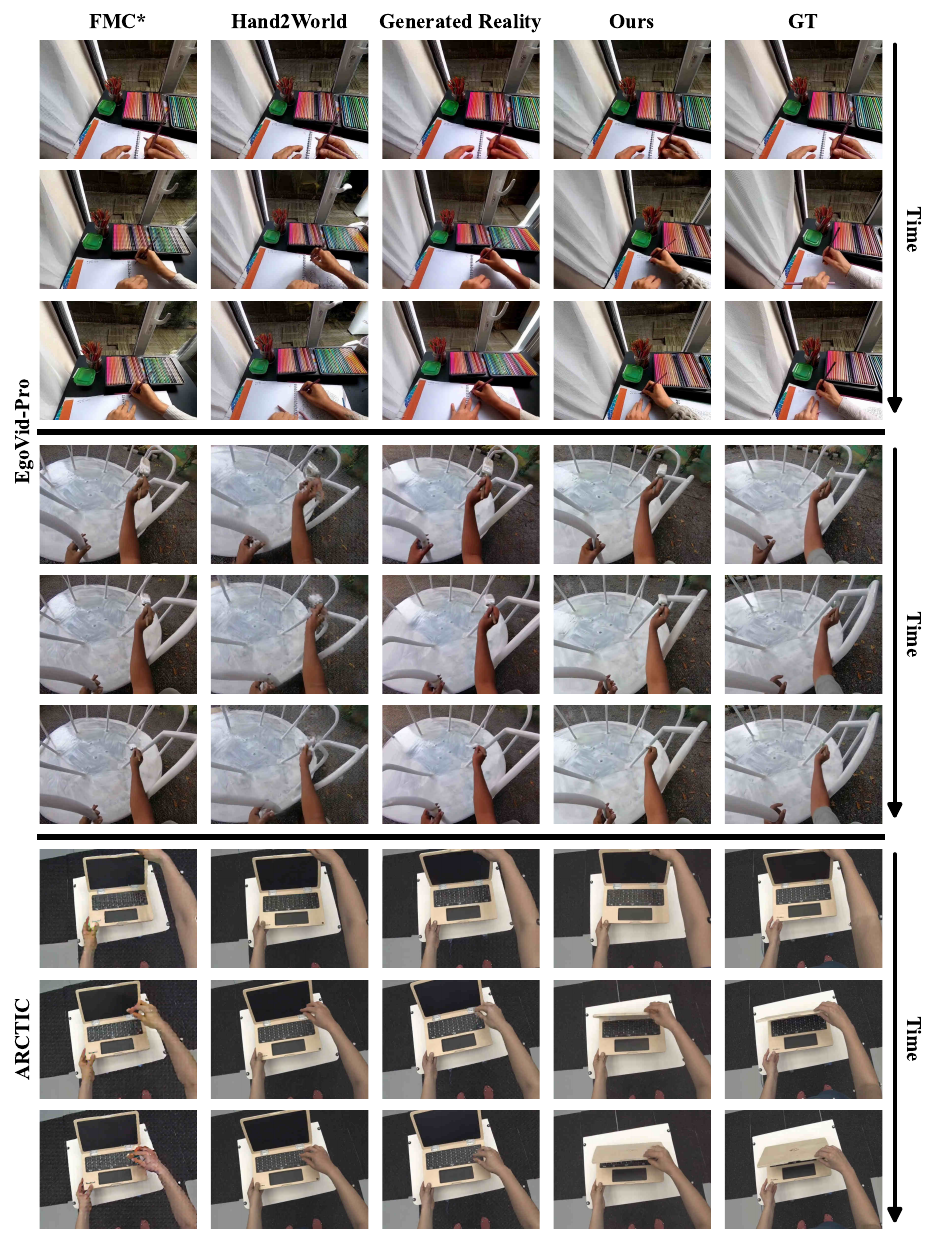}
  \caption{\textbf{Additional qualitative comparison against baseline
    methods.} Extending \cref{fig:compare-method} with more examples.}
  \label{fig:supp_more_method}
\end{figure}

\paragraph{More qualitative results.}
We provide additional qualitative comparisons to complement those in the
main paper.
\Cref{fig:supp_more_data} shows further examples across different
training-data conditions, and \cref{fig:supp_more_method} shows further
comparisons against the baseline methods.
Across these additional examples, which span a wide range of everyday
scenes, general hand-object interactions, and camera ego-motion patterns, our
method consistently produces the most realistic hand interactions and the
most accurate adherence to the conditioning hand trajectory.
The camera-space baselines repeatedly entangle the camera trajectory with
the hand articulation, misplacing the hand or missing the intended contact,
whereas our world-space Pl\"{u}cker-ray representation keeps the hand
geometrically consistent even under rapid camera motion.
These qualitative trends hold uniformly across scenarios and corroborate the
quantitative advantages reported in the main paper, underscoring the
robustness and generality of our approach.

\section{More Ablation Studies}

\paragraph{World-space representations.}
To isolate the effectiveness of our Pl\"{u}cker representation, we ablate
against three world-space alternatives that share the same hand-mesh
rasterization but encode different per-pixel attributes: a \emph{position
map} (world-space $xyz$), a \emph{depth map} (scalar distance along the camera
ray), and a \emph{normal map} (3-channel world-space surface normal). All
variants use the same backbone and EgoVid-Pro training data and differ only in
the attributes encoded on the hand silhouette.
\Cref{tab:supp_world_signal_ablation} reports five hand-pose metrics on
EgoVid-Pro. Our Pl\"{u}cker line jointly encodes surface orientation and
absolute world-space placement in 6D, subsuming cues captured only partially
by the alternatives and yielding the most accurate hand control.

\begin{table*}[t]
  \centering
  \caption{Ablation of the geometric guidance scale $w_{\mathrm{geo}}$
    across two training-data configurations.
    The text guidance scale is fixed at $w_{\mathrm{text}}=1$.
    \textbf{Bold} marks the best per column; \underline{underline} marks
    the second-best.}
  \label{tab:supp_cfg_geo}
  \tabcolsep=2pt
  \small
  \begin{tabular}{llccccccccccc}
    \toprule
    & & \multicolumn{6}{c}{Visual Quality} & \multicolumn{2}{c}{Camera Pose} & \multicolumn{3}{c}{Hand Pose}\\
    \cmidrule(lr){3-8}\cmidrule(lr){9-10}\cmidrule(lr){11-13}
    Training Data & $w_{\mathrm{geo}}$ & PSNR$\uparrow$ & SSIM$\uparrow$ & LPIPS$\downarrow$ & FVD$\downarrow$ & Sub.Con.$\uparrow$ & BG.Con.$\uparrow$ & RotErr$\downarrow$ & TransErr$\downarrow$ & L2Err$\downarrow$ & PA-JPE$\downarrow$ & Recall$\uparrow$ \\
    \midrule
    \multirow{2}{*}{ARCTIC + EgoVid (raw)}
      & 1 & 16.05 & 0.4835 & 0.3759 & 517.34 & 0.9109 & 0.9015 & 5.07          & 5.35          & 16.14          & \underline{7.36}          & 90.30 \\
      & 2 & 14.87 & 0.4581 & 0.3997 & 550.20 & 0.9033 & 0.8938 & \underline{4.03} & 4.91       & \underline{13.08} & 7.56 & \underline{91.85} \\
    \midrule
    \multirow{2}{*}{\textbf{Ours (EgoVid-Pro)}}
      & 1 & \underline{17.21} & \underline{0.5168} & \underline{0.3374} & \underline{357.65} & \underline{0.9168} & \textbf{0.9226}    & 5.06          & \underline{4.66} & 21.91          & 9.58          & 83.55 \\
      & 2 & \textbf{17.39}    & \textbf{0.5354}    & \textbf{0.2943}    & \textbf{266.82}    & \textbf{0.9305}    & \underline{0.9224} & \textbf{3.73} & \textbf{3.35}    & \textbf{11.83} & \textbf{6.43} & \textbf{94.62} \\
    \bottomrule
  \end{tabular}%
\end{table*}

\paragraph{Text CFG Guidance Scale.}
We sweep the text guidance scale $w_{\mathrm{text}}$ while fixing the
geometric guidance scale at $w_{\mathrm{geo}}=2$ and report the same
metrics used in the main paper's data-quality comparison
(\cref{tab:data_effectiveness}).
\Cref{tab:supp_cfg_text} shows that no single $w_{\mathrm{text}}$
dominates: $w_{\mathrm{text}}=2$ achieves the strongest visual quality
, $w_{\mathrm{text}}=3$ is best on most
hand-pose metrics, and $w_{\mathrm{text}}=1$
yields the lowest FVD and camera rotation error.
We adopt $w_{\mathrm{text}}=3$ as our default for its balanced
visual-quality profile.

\paragraph{Geometric CFG Guidance Scale.}
We further ablate the geometric guidance scale $w_{\mathrm{geo}}$ on two
training-data configurations: raw EgoVid clips combined with ARCTIC, and
our curated EgoVid-Pro.
The text guidance scale is fixed at $w_{\mathrm{text}}=1$ for both
configurations.

\Cref{tab:supp_cfg_geo} shows that for both configurations,
increasing $w_{\mathrm{geo}}$ improves the control-related metrics, confirming that stronger geometric guidance enforces tighter
alignment between the generated video and the conditioning hand
trajectory.
However, the two configurations diverge sharply on visual quality.
For the raw-data variant, stronger guidance degrades the overall visual qualities, indicating that the raw annotations conflict with the
generator's image prior, pushing the model harder to satisfy them comes at
the cost of image fidelity.
Our annotated EgoVid-Pro removes this conflict.
Increasing $w_{\mathrm{geo}}$ on our model simultaneously improves
not only the visual-quality metrics but also the hand-pose metrics: the
control signal and the image prior agree.
This decoupling between geometric control fidelity and visual quality is a
direct consequence of the clean, protagonist-centered annotations
produced by our pipeline.

\begin{table}[t]
  \centering
  \caption{Ablation study on world-space hand control signals on EgoVid-Pro.
    \textbf{Bold} and \underline{underline} mark the best and second-best
    values per column, respectively.}
  \label{tab:supp_world_signal_ablation}
  \setlength{\tabcolsep}{3.5pt}
  \begin{tabular}{lccccc}
    \toprule
    Signal & L2Err$\downarrow$ & PA-JPE$\downarrow$ & W-JPE$\downarrow$ & WA-JPE$\downarrow$ & RTE$\downarrow$ \\
    \midrule
    Position      & 16.10             & 7.08             & 71.46             & 37.94             & 4.98 \\
    Depth         & 13.90             & \underline{6.60} & 76.81             & 35.67             & 4.98 \\
    Normal        & \underline{12.37} & 6.63             & \underline{69.24} & \underline{34.26} & \underline{4.49} \\
    \textbf{Ours} & \textbf{11.32}    & \textbf{6.37}    & \textbf{67.46}    & \textbf{33.02}    & \textbf{4.12} \\
    \bottomrule
  \end{tabular}
\end{table}

\section{Limitations and Future Work}

\paragraph{Occlusion.}
Our control signal is produced by rasterizing the visible hand surface, so
it does not explicitly encode pose changes in self-occluded hand regions.
Layered representations, such as MPI-style encodings, may help expose or
preserve these hidden hand states.

\paragraph{Missing detections.}
Our annotation pipeline relies on an off-the-shelf hand detection and tracking tool.
When a frame has no detection, the system cannot reliably distinguish whether
the hand has left the image or the detector simply missed it.
To protect video-generation quality, we currently avoid using clips with such
uncertain long gaps rather than hallucinating supervision for them.
Mask-based training or uncertainty-aware conditioning may make it possible to
learn from these ambiguous segments in future work.

\end{document}